%% file: main.tex
\documentclass[letterpaper, 10 pt, journal, twoside]{IEEEtran}

\IEEEoverridecommandlockouts                              

\usepackage{amsmath}
\usepackage{amssymb}
\usepackage{amsthm} 
\usepackage{mathtools}
\usepackage[hidelinks]{hyperref}
\usepackage[capitalize]{cleveref}
\usepackage{xspace}
\usepackage{graphicx}
\usepackage[dvipsnames]{xcolor}
\usepackage{pgfplots}
\usepgfplotslibrary{groupplots}
\pgfplotsset{compat=1.9}
\usepackage{booktabs, nicematrix}
\usepackage{adjustbox}
\usepackage{xfrac}

\allowdisplaybreaks

\usepackage{ifthen}
\newboolean{anonymous}
\setboolean{anonymous}{false}
\newboolean{revisions}
\setboolean{revisions}{false}
\newboolean{finalrevisions}
\setboolean{finalrevisions}{false}
\newboolean{archive}
\setboolean{archive}{true}

\input{preamble}

\input{preamble_diagram}

\title{
GRAND: Guidance, Rebalancing, and Assignment for Networked Dispatch in Multi-Agent Path Finding
}

\ifthenelse{\boolean{anonymous}}{%
  \author{Anonymous Authors}%
}{
\author{Johannes Gaber$^{1,*}$, Meshal Alharbi$^{1,*}$, Daniele Gammelli$^{2}$, 
        and Gioele Zardini${^1}$
\arxiv{\thanks{Manuscript received: December 2, 2025; Revised February 6, 2026; Accepted March 3, 2026.}}
\arxiv{\thanks{This paper was recommended for publication by Editor Aniket Bera upon evaluation of the Associate Editor and Reviewers’ comments.}}
\thanks{This work was supported by Prof. Zardini's grant from the MIT Amazon Science Hub, hosted in the MIT Schwarzman College of Computing, and the MIT Maritime Consortium.
This material is based upon work supported by the Defense Advanced Research Projects Agency (DARPA) under Award No. D25AC00373. The views and conclusions contained in this document are those of the authors and should not be interpreted as representing the official policies, either expressed or implied, of the U.S. Government.
}
\thanks{$^1$Johannes Gaber, Meshal Alharbi, and Gioele Zardini are with the Laboratory for Information and Decision Systems, Massachusetts Institute of Technology, Cambridge, MA, USA (e-mails: {\tt\footnotesize \{jgaber, meshal, gzardini\}@mit.edu}).}
\thanks{$^2$Daniele Gammelli is with the Department of Aeronautics and Astronautics, Stanford University (e-mail: {\tt\footnotesize gammelli@stanford.edu}).}
\thanks{$^*$Gaber and Alharbi contributed equally to this work.}
\arxiv{\thanks{Digital Object Identifier (DOI): see top of this page.}}
}
}

\arxiv{
\markboth{IEEE ROBOTICS AND AUTOMATION LETTERS. PREPRINT VERSION. ACCEPTED MARCH, 2026}
{GABER \MakeLowercase{\textit{et al.}}: GUIDANCE, REBALANCING, AND ASSIGNMENT FOR NETWORKED DISPATCH IN MULTI-AGENT PATH FINDING}
}

\begin{document}

\maketitle


\input{sections/01-abstract}
\arxiv{
}
\input{sections/02-introduction}
\input{sections/03-related-work}
\input{sections/04-background}
\input{sections/05-problem-formulation}
\input{sections/06-method}
\input{sections/07-results}

\input{sections/08-conclusion}

\bibliographystyle{IEEEtran}
\bibliography{references}


\end{document}

%% file: preamble.tex
\DeclareRobustCommand{\rev}[1]{%
  \ifthenelse{\boolean{revisions}}{%
    \ifmmode \begingroup\color{blue}#1\endgroup
    \else   \textcolor{blue}{#1}%
    \fi
  }{#1}%
}
\DeclareRobustCommand{\frev}[1]{%
  \ifthenelse{\boolean{finalrevisions}}{%
    \ifmmode \begingroup\color{blue}#1\endgroup
    \else   \textcolor{blue}{#1}%
    \fi
  }{#1}%
}
\DeclareRobustCommand{\arxiv}[1]{%
  \ifthenelse{\boolean{archive}}{}{#1}%
}
\theoremstyle{definition}
\newtheorem{definition}{Definition}[section]

\newtheorem{remark}{Remark}[section]

\usepackage[acronym]{glossaries-extra}
\setabbreviationstyle[acronym]{long-short}
\glssetcategoryattribute{acronym}{nohyperfirst}{true}

\newacronym{acr:rl}{RL}{Reinforcement Learning}
\newacronym{acr:sac}{SAC}{Soft Actor Critic}
\newacronym{acr:mapf}{MAPF}{Multi-Agent Path-Finding}
\newacronym{acr:lmapf}{LMAPF}{Lifelong Multi-Agent Path-Finding}
\newacronym{acr:ts}{TS}{Task Scheduling}
\newacronym{acr:lts}{LTS}{Lifelong Task Scheduling}
\newacronym{acr:mapd}{MAPD}{Multi-Agent Pickup and Delivery}
\newacronym{acr:ilp}{ILP}{Integer Linear Program}
\newacronym{acr:lrr}{\rev{LoRR}}{\rev{League of Robot Runners}}
\newacronym{acr:wppl}{WPPL}{Windowed Parallel PIBT-LNS}
\newacronym{acr:tfgp}{TFGP}{Traffic-Flow Guided PIBT}
\newacronym{acr:gnn}{GNN}{Graph Neural Network}
\newacronym{acr:mdp}{MDP}{Markov Decision Process}

\newcommand{\ALGLRR}{\textsc{\rev{LoRR} Winner}\xspace}
\newcommand{\ALGILP}{\textsc{G-OPT}\xspace}
\newcommand{\ALGG}{\textsc{Greedy}\xspace}
\newcommand{\ALGRL}{\textsc{\rev{GRAND}}\xspace}

\newcommand{\Scale}{0.90}
\newcommand{\TableVerStretch}{1.1}

\pgfplotsset{
  styleGreedy/.style={fill=Apricot!80, draw=Apricot!80!black, line width=0.2pt},
  styleLRR/.style={fill=Red!80, draw=Red!80!black, line width=0.2pt},
  styleLRR2/.style={color=Red!80},
  styleRL/.style={fill=RoyalBlue!80, draw=RoyalBlue!80!black, line width=0.2pt},
  styleRL2/.style={color=RoyalBlue!80},
  styleILP/.style={fill=ForestGreen!80, draw=ForestGreen!80!black, line width=0.2pt}, 
}

\newcommand{\tup}[1]{\left( #1\right)}

%% file: preamble_diagram.tex
\usepackage{tikz}
\usetikzlibrary{shapes.geometric, arrows.meta, positioning, calc, fit, backgrounds, shadows.blur, patterns, 3d}

\definecolor{guidanceColor}{RGB}{235, 245, 251} 
\definecolor{guidanceStroke}{RGB}{52, 152, 219}
\definecolor{rebalColor}{RGB}{254, 249, 231}    
\definecolor{rebalStroke}{RGB}{241, 196, 15}
\definecolor{matchColor}{RGB}{232, 248, 245}    
\definecolor{matchStroke}{RGB}{26, 188, 156}
\definecolor{darkgray}{RGB}{80, 80, 80}

\newcommand{\drawThreeDBar}[4]{
    \pgfmathsetmacro{\xla}{#1-0.2} \pgfmathsetmacro{\xra}{#1+0.2}
    \pgfmathsetmacro{\yla}{#2-0.2} \pgfmathsetmacro{\yra}{#2+0.2}
    \pgfmathsetmacro{\zh}{#3}

    \fill[#4!60] (\xla,\yla,0) -- (\xra,\yla,0) -- (\xra,\yla,\zh) -- (\xla,\yla,\zh) -- cycle;
    \fill[#4!80] (\xra,\yla,0) -- (\xra,\yra,0) -- (\xra,\yra,\zh) -- (\xra,\yla,\zh) -- cycle;
    \fill[#4!40] (\xla,\yla,\zh) -- (\xra,\yla,\zh) -- (\xra,\yra,\zh) -- (\xla,\yra,\zh) -- cycle;
    \draw[#4!80, very thin] (\xla,\yla,\zh) -- (\xra,\yla,\zh) -- (\xra,\yra,\zh) -- (\xla,\yra,\zh) -- cycle;
    \draw[#4!80, very thin] (\xra,\yla,0) -- (\xra,\yla,\zh);
    \draw[#4!80, very thin] (\xla,\yla,0) -- (\xla,\yla,\zh);
    \draw[#4!80, very thin] (\xra,\yra,0) -- (\xra,\yra,\zh);
}

\makeatletter
\newcommand\resetstackedplots{%
  \pgfplots@stacked@isfirstplottrue
  \addplot[forget plot, draw=none] coordinates {
  ({$|A|\,{=}\,100$\\$|V_\mathrm{tile}|\,{=}\,598 $}, 0)
  ({$|A|\,{=}\,200$\\$|V_\mathrm{tile}|\,{=}\,975 $}, 0)
  ({$|A|\,{=}\,300$\\$|V_\mathrm{tile}|\,{=}\,1333$}, 0)
  ({$|A|\,{=}\,500$\\$|V_\mathrm{tile}|\,{=}\,2537$}, 0)
};
}
\makeatother

%% file: sections/01-abstract.tex
\begin{abstract}
Large robot fleets are now common in warehouses and other logistics settings, where small control gains translate into large operational impacts.
In this article, we address task scheduling for lifelong Multi-Agent Pickup-and-Delivery (MAPD) and propose a hybrid method that couples learning-based global guidance with lightweight optimization.
A graph neural network policy trained via reinforcement learning outputs a desired distribution of free agents over an aggregated warehouse graph.
This signal is converted into region-to-region rebalancing through a minimum-cost flow, and finalized by small, local assignment problems, preserving accuracy while keeping per-step latency within a 1\,s compute budget.
\frev{We call this approach GRAND: a hierarchical algorithm that relies on Guidance, Rebalancing, and Assignment to explicitly leverage the workspace Network structure and Dispatch agents to tasks.}
On congested warehouse benchmarks from the \rev{League of Robot Runners (LoRR)} with up to 500 agents, our approach improves throughput by up to 10\% over the 2024 winning scheduler while maintaining real-time execution.
The results indicate that coupling graph-structured learned guidance with tractable solvers reduces congestion and yields a practical, scalable blueprint for high-throughput scheduling in large fleets.
\end{abstract}

%% file: sections/02-introduction.tex
\section{Introduction}
\label{sec:intro}

\IEEEPARstart{L}{arge}
fleets of mobile robots are now common in domains such as autonomous ride-hailing~\cite{zardini2022analysis} and warehouse automation~\cite{d2012guest}.
At scale, even small coordination gains translate into substantial economic and environmental impacts.
For instance, city-scale robotaxi deployments and million-robot fulfillment centers underscore the opportunity for better fleet control~\cite{waymo2025, amazon2025}. 
In these systems, a scheduler assigns robots to tasks (i.e., \gls{acr:ts}) and a planner generates collision-free motions (i.e., \gls{acr:mapf}). 
These two problems are sometimes called task planning and motion planning, respectively.
The warehouse and ride-hailing settings are well captured by \gls{acr:mapd}, where each task requires visiting two locations in order~\cite{ma2017lifelong}. 
The lifelong variants \gls{acr:lts} and \gls{acr:lmapf} model continuous task arrivals. 
Both problems are known to be NP-hard~\cite{yu2013structure,doring2025parameterized}, making exact, monolithic optimization impractical at realistic scales and control rates.

In this context, three families of approaches are prevalent \cite{antonyshyn2023multiple}.
First, optimization-based schedulers (e.g., \gls{acr:ilp}/Hungarian method on A$^\star$ distances) provide clean models but can be myopic to congestion and expensive at large scale. 
Second, heuristics and rules are fast and widely used in practice, 
yet may sacrifice throughput under heavy coupling.
Finally, learning-based methods promise fast inference and the ability to exploit dynamics beyond simplified models, but often lack guarantees and have
not consistently outperformed strong heuristics in classic \gls{acr:mapf} settings.

\input{figures/overview_diagram}

\paragraph*{Statement of Contribution}
In this paper, we present a hybrid \gls{acr:lts} architecture that cleanly separates \emph{learned global guidance} from \emph{lightweight combinatorial assignment}. 
A \gls{acr:gnn} policy trained with \gls{acr:rl} outputs a desired distribution of free agents over an aggregated warehouse graph; this signal is converted into region-to-region rebalancing via a minimum-cost flow formulation and completed with small, local \glspl{acr:ilp}, preserving assignment accuracy while keeping per-step latency within a $1$\,s control budget.
\rev{We call this approach GRAND: a hierarchical algorithm that relies on Guidance, Rebalancing, and Assignment to explicitly leverage the workspace Network structure and Dispatch agents to tasks.}
\cref{fig:overview_diagram} showcases an overview diagram of our methodology.
On congested warehouse benchmarks from the \gls{acr:lrr}~\cite{chan2024league}, the resulting scheduler improves throughput by up to $10\%$ over the 2024 winning baseline, demonstrating that coupling graph-structured learned guidance with tractable solvers yields robust, high-throughput control for large fleets.

%% file: figures/overview_diagram.tex
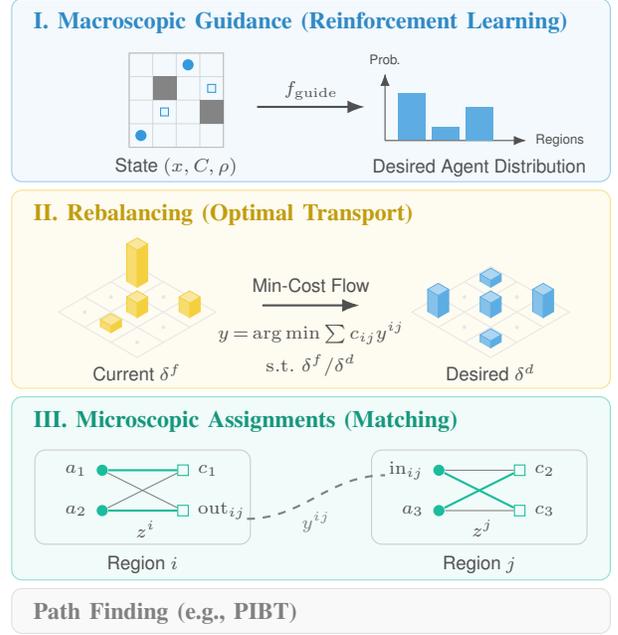
\begin{figure}[t]
\centering
\setlength{\abovecaptionskip}{2pt}
\begin{adjustbox}{max width=0.9\columnwidth, center} 
\begin{tikzpicture}[
    font=\sffamily\scriptsize,
    text=darkgray,
    >=LaTeX,
    x=1cm, y=1cm,
    iso/.style={x={(0.866cm,-0.5cm)}, y={(0.866cm,0.5cm)}, z={(0cm,1cm)}},
    scale=0.9,
]

    
    \node[rounded corners, fill=guidanceColor!60, draw=guidanceStroke!50, minimum width=8cm, minimum height=2.45cm] (box1) at (0,0) {};
    \node[anchor=north west, text=guidanceStroke!90!black, font=\bfseries\small] at (box1.north west) [xshift=5pt, yshift=-2pt] {I. Macroscopic Guidance (Reinforcement Learning)};

    \begin{scope}[shift={(-2.0, -0.15)}, scale=0.7]
        \draw[step=0.5, gray!30, thin] (-1,-1) grid (1,1);
        \draw[gray!50, thin] (-1,-1) rectangle (1,1);
        
        \fill[darkgray!70] (-0.5, 0) rectangle (0, 0.5);
        \fill[darkgray!70] (0.5, -0.5) rectangle (1, 0);

        \node[circle, fill=guidanceStroke, inner sep=1.5pt] at (-0.75, -0.75) {};
        \node[circle, fill=guidanceStroke, inner sep=1.5pt] at (0.25, 0.75) {};

        \node[draw=guidanceStroke, fill=guidanceColor, rectangle, inner sep=1.5pt] at (0.75, 0.25) {};
        \node[draw=guidanceStroke, fill=guidanceColor, rectangle, inner sep=1.5pt] at (-0.25, -0.25) {};

        \node[below=0.65cm, darkgray, align=center] at (0,0) {State $(x, C, \rho)$};
    \end{scope}

    \node[font=\sffamily\scriptsize, align=center] at (0, 0) {$f_\mathrm{guide}$};
    \draw[->, thick, darkgray] (-0.8, -0.25) -- (0.8, -0.25);

    \begin{scope}[shift={(2.6, -0.75)}]
        \draw[->, darkgray, font=\sffamily\tiny] (-1.5, 0) -- (0.6, 0) node[right] {Regions};
        \draw[->, darkgray, font=\sffamily\tiny] (-1.5, 0) -- (-1.5, 1) node[above] {Prob.};
        \draw[fill=guidanceStroke!80, draw=none] (-1.3, 0.01) rectangle (-0.9, 0.7);
        \draw[fill=guidanceStroke!80, draw=none] (-0.8, 0.01) rectangle (-0.4, 0.2);
        \draw[fill=guidanceStroke!80, draw=none] (-0.3, 0.01) rectangle (0.1, 0.5);
        \node[align=center] at (-0.1, -0.4) {Desired Agent Distribution};
    \end{scope}

    
    \node[rounded corners, fill=rebalColor!60, draw=rebalStroke!50, minimum width=8cm, minimum height=2.65cm, below=0.1cm of box1] (box2) {};
    \node[anchor=north west, text=rebalStroke!80!black, font=\bfseries\small] at (box2.north west) [xshift=5pt, yshift=-2pt] {II. Rebalancing (Optimal Transport)};

    \begin{scope}[shift={(-3.75, -3.3)}, scale=0.45, iso]
        \draw[gray!20] (0,0,0) -- (3,0,0) -- (3,3,0) -- (0,3,0) -- cycle;
        \foreach \x in {1,2} \draw[gray!20] (\x,0,0) -- (\x,3,0);
        \foreach \y in {1,2} \draw[gray!20] (0,\y,0) -- (3,\y,0);
        \foreach \x in {0.5, 1.5, 2.5} \foreach \y in {0.5, 1.5, 2.5}
            \node[circle, fill=gray!20, inner sep=0.5pt] at (\x,\y,0) {};

        \drawThreeDBar{0.5}{2.5}{1.25}{rebalStroke}
        %
        %
        \drawThreeDBar{1.5}{1.5}{0.50}{rebalStroke}
        \drawThreeDBar{2.5}{2.5}{0.50}{rebalStroke}
        \drawThreeDBar{1.5}{0.5}{0.25}{rebalStroke}
        %

        \node[below, darkgray, font=\sffamily\scriptsize] at (3, 0, 0.1) {Current $\delta^f$};
    \end{scope}

    \node[font=\sffamily\scriptsize, align=center] at (0, -2.9) {Min-Cost Flow};
    \draw[->, thick, darkgray, shorten >= 2pt, shorten <= 2pt] (-0.8, -3.2) -- (0.8, -3.2);
    \node[font=\scriptsize, align=center] at (0, -3.85) {$y\,{=}\arg\min \sum c_{ij} y^{ij}$ \\[4pt] $\mathrm{s.t.}\ \delta^f / \delta^d$};

    \begin{scope}[shift={(1.50, -3.3)}, scale=0.45, iso]
        \draw[gray!20] (0,0,0) -- (3,0,0) -- (3,3,0) -- (0,3,0) -- cycle;
        \foreach \x in {1,2} \draw[gray!20] (\x,0,0) -- (\x,3,0);
        \foreach \y in {1,2} \draw[gray!20] (0,\y,0) -- (3,\y,0);
        \foreach \x in {0.5, 1.5, 2.5} \foreach \y in {0.5, 1.5, 2.5}
            \node[circle, fill=gray!20, inner sep=0.5pt] at (\x,\y,0) {};

        \drawThreeDBar{0.5}{2.5}{0.25}{guidanceStroke}
        %
        %
        \drawThreeDBar{0.5}{0.5}{0.75}{guidanceStroke}
        \drawThreeDBar{1.5}{1.5}{0.50}{guidanceStroke}
        \drawThreeDBar{2.5}{2.5}{0.75}{guidanceStroke}
        %
        %
        \drawThreeDBar{2.5}{0.5}{0.25}{guidanceStroke}

        \node[below, darkgray, font=\sffamily\scriptsize] at (3, 0, 0.1) {Desired $\delta^d$};
    \end{scope}


    \node[rounded corners, fill=matchColor!60, draw=matchStroke!50, minimum width=8cm, minimum height=2.45cm, below=0.1cm of box2] (box3) {};
    \node[anchor=north west, text=matchStroke!80!black, font=\bfseries\small] at (box3.north west) [xshift=5pt, yshift=-2pt] {III. Microscopic Assignments (Matching)};

    \begin{scope}[shift={(-2.5, -5.55)}] 
        \node[text=darkgray] at (0, -1.5) {Region $i$};
        \draw[gray!40, rounded corners] (-1.6, -1.2) rectangle (1.6, 0.2);
        \node[circle, fill=matchStroke, inner sep=1.5pt, label=left:\scriptsize $a_1$] (a1) at (-0.6, -0.1) {};
        \node[circle, fill=matchStroke, inner sep=1.5pt, label=left:\scriptsize $a_2$] (a2) at (-0.6, -0.7) {};
        \node[draw=matchStroke, rectangle, inner sep=2pt, label=right:\scriptsize $c_1$] (t1) at (0.6, -0.1) {};
        \node[draw=matchStroke, rectangle, inner sep=2pt, label=right:\scriptsize $\mathrm{out}_{ij}$] (t2) at (0.6, -0.7) {};
        \draw[thick, matchStroke] (a1) -- (t1);
        \draw[gray] (a1) -- (t2);
        \draw[gray] (a2) -- (t1);
        \draw[thick, matchStroke] (a2) -- (t2) node[below, darkgray, xshift=-0.5cm] {$z^i$};
    \end{scope}

    \begin{scope}[shift={(2.5, -5.55)}] 
        \node[text=darkgray] at (0, -1.5) {Region $j$};
        \draw[gray!40, rounded corners] (-1.6,-1.2) rectangle (1.6, 0.2);
        \node[circle, fill=matchStroke, inner sep=1.5pt, label=left:\scriptsize $\mathrm{in}_{ij}$] (a3) at (-0.6, -0.1) {};
        \node[circle, fill=matchStroke, inner sep=1.5pt, label=left:\scriptsize $a_3$] (a4) at (-0.6, -0.7) {};
        \node[draw=matchStroke, rectangle, inner sep=2pt, label=right:\scriptsize $c_2$] (t3) at (0.6, -0.1) {};
        \node[draw=matchStroke, rectangle, inner sep=2pt, label=right:\scriptsize $c_3$] (t4) at (0.6, -0.7) {};
        \draw[gray] (a3) -- (t3);
        \draw[thick, matchStroke] (a3) -- (t4);
        \draw[thick, matchStroke] (a4) -- (t3);
        \draw[gray] (a4) -- (t4) node[below, darkgray, xshift=-0.5cm] {$z^j$};
    \end{scope}

    \draw[thick, gray, dashed] (-0.95, -6.3725) to[out=0, in=180] node[below, yshift=-2pt] {$y^{ij}$} (1.10, -5.725);


    \node[rounded corners, fill=gray!5, draw=gray!25,, minimum width=8cm, minimum height=0.60cm, below=0.1cm of box3] (box4) {};
    \node[anchor=north west, text=gray, font=\bfseries\small] at (box4.north west) [xshift=5pt, yshift=-2pt] {Path Finding (e.g., PIBT)};

\end{tikzpicture}
\end{adjustbox}
\caption{
\rev{Overview of GRAND, our hierarchical task scheduling approach}: 
(I) a data-driven layer provides global guidance as a desired agent distribution;
(II) a region-to-region optimal transport rebalances free agents toward the desired distribution; 
and (III) local, decoupled matching problems produce the final assignments for task scheduling.
All symbols are defined later in the text.
}
\label{fig:overview_diagram}
\end{figure}

%% file: sections/03-related-work.tex
\section{Related Work}
\label{sec:literature}

\gls{acr:ts} in \gls{acr:mapf}/\gls{acr:mapd} complements collision-aware planning and has been explored mainly via distance-based assignment, heuristics, and learning \cite{antonyshyn2023multiple}.
Ignoring path interactions, \gls{acr:ts} reduces to a
weighted bipartite matching between free agents and tasks using A$^\star$-based distances~\cite{hart1968formal}.
The assignment can be solved via \gls{acr:ilp} or the Hungarian method~\cite{ramshaw2012minimum}, and appears in early \gls{acr:mapd} formulations coupling online re-assignment with matching~\cite{ma2017lifelong}.
In congested warehouses, however, distance-optimal allocations at dispatch time need not remain efficient during execution due to conflicts and queuing, motivating faster rules and metaheuristics. 
For instance, large neighborhood search has been applied to multi-goal \gls{acr:mapd}~\cite{xu2022multi}. 
The \gls{acr:lrr}~\cite{chan2024league} simulator was recently introduced to standardize comparisons for \gls{acr:mapd}, providing warehouse-style benchmarks among others. The \gls{acr:lrr} has spurred congestion-aware greedy strategies that perform strongly in practice, including a runner-up based on congestion prediction~\cite{gao2025adaptive} and the winning 2024 scheduler leveraging refined greedy priorities~\cite{yukhnevich2025enhancing}.

Beyond heuristics, a flow-based model that computes a global, system-wide matching between agents and tasks was proposed to guide assignment in sortation centers~\cite{kou2020idle}, and several works couple assignment with planning. Chen et al.\ integrate \gls{acr:ts} and path planning for capacitated \gls{acr:mapd} in a unified framework~\cite{chen2021integrated}.
Newer variants expand problem scope, e.g., online \gls{acr:mapd} with deadlines (\gls{acr:mapd}-D) and dynamic path finding for multi-load agents~\cite{makino2024online}.
System-level closed-loop formulations that unify scheduling and planning are also emerging (e.g., FICO~\cite{li2025fico}). 
At scale, the \gls{acr:lmapf} setting highlights the need for high-quality decisions under tight control budgets and heavy congestion~\cite{chan2024league}.

Learning-based \gls{acr:ts} and \gls{acr:lts} for warehouse \gls{acr:mapd} are comparatively sparse~\cite{wang2025breaking}, especially when compared to the learning-based literature for \gls{acr:mapf}~\cite{wang2025paths}.
Agrawal et al.\ employ attention-based \gls{acr:rl} to directly allocate tasks sequentially~\cite{agrawal2023rtaw}. 
Related large-fleet dispatch domains provide complementary ideas: high-capacity ride-pooling uses graph-structured \gls{acr:ilp} to jointly assign requests and vehicles~\cite{alonso2017demand}, while \gls{acr:gnn} \gls{acr:rl} has been used to rebalance fleets and improve downstream matching~\cite{gammelli2021graph, gammelli2023graph, tresca2025robo}. 
Our approach follows this hybrid trajectory: learned global guidance on a graph shapes downstream optimization (flow/matching), aiming to capture congestion effects while retaining the reliability and precision of combinatorial solvers, and we evaluate in \gls{acr:lrr}’s standardized warehouse environment~\cite{chan2024league}.

%% file: sections/04-background.tex
\section{Background}
\label{sec:background}

\subsection{Reinforcement Learning}
\label{sec:rl_background}

\gls{acr:rl} is a framework for solving sequential decision-making problems that are often modeled as a fully observed \gls{acr:mdp}. An \gls{acr:mdp} is characterized by a tuple $\mathcal{M}=\tup{\mathcal{S}, \mathcal{A},P,d_0,r,\gamma}$, where $\mathcal{S}$ and $\mathcal{A}$ denote the state and action spaces.
$P(s_{t+1} \mid s_t,a_t)$ is the transition kernel, $d_0$ is the initial state distribution, $r: \mathcal{S} \times \mathcal{A} \to \mathbb{R}$ is the reward function, and $\gamma\in(0,1]$ is the discount factor. A (stochastic) policy $\pi(a \mid s)$ maps states to a distribution over actions. For a finite horizon $H$, a trajectory is
$\tau=(s_0,a_0,\ldots,s_{H-1},a_{H-1},s_H)$, and the probability $p_\pi(\tau)$ of this trajectory under policy $\pi$ is
$p_\pi(\tau) = d_0(s_0) \prod_{t=0}^{H-1} \pi(a_t\mid s_t)\,P(s_{t+1}\mid s_t,a_t).$
%
%
The \gls{acr:rl} objective is to find a policy that maximizes the discounted return:
$J(\pi) = \mathbb{E}_{\tau\sim p_\pi} \!\big[ \sum_{t=0}^{H-1} \gamma^{t}\,r(s_t,a_t) \big].$
%
%
Data typically come from interacting with the environment using some behavior policy $\beta$, often the current version of $\pi$ under training. Each interaction yields a transition $(s_t,a_t,r_t,s_{t+1})$, where $r_t = r(s_t,a_t)$, which is used to 
improve the policy.

%% file: sections/05-problem-formulation.tex
\section{Problem Formulation}
\label{sec:formulation}

We study \gls{acr:lts} for multiple agents~$A=\{a_1,\dots,a_N\}$ moving on a directed, reflexive graph~$G=\tup{V,E}$ over discrete timesteps~$t=0,1,\dots,T_{\max}$. 
Reflexivity encodes the option to wait via self-loops~$\tup{v,v}\in E$.
At each timestep, a scheduler assigns per-agent goals, a planner generates collision-free motions, and a task generator updates completed tasks with new ones.
Let~$x_t\colon A\to V$ denote the joint position of all agents at time~$t$ (for convenience,~$v_t^{a}\coloneqq x_t(a)$), and let~$C_t\subseteq V$ denote the set of uncompleted tasks at time~$t$.
We consider a lifelong setting in which the number of available tasks is constant:~$\vert C_t\vert=M\in \mathbb{N}$ for all~$t$.


\subsection{Mathematical Model}
\label{sec:math_model}

We view~$\tup{x_t, C_t}$ as the system state.
Its evolution is governed by the task generator, the scheduling policy, and the planning policy.
\begin{definition}[Task generator]
A \emph{task generator} is a map~$f_{\mathrm{TG}}: V^A \times 2^V \to 2^V$ that produces the next task set~$C_{t+1} = f_{\mathrm{TG}} \tup{x_t, C_t}.$
\rev{If an agent~$a$} completes a task at time~$t$ (i.e.,~$x_t(a)\in C_t$), that task is removed in~$C_{t+1}$ and immediately replaced by a new task.
All other tasks carry over from~$C_t$ to~$C_{t+1}$.
This maintains~$\vert C_{t+1}\vert=M$.
\end{definition}
In \gls{acr:mapd}, a task may comprise an ordered sequence of subgoals completed by a single agent.
Now, we formalize our notion of a scheduling policy. 
\begin{definition}[Goal map]
A \emph{goal map} at time $t$ is an injective map $\rho_t : A \to V$.
\end{definition}
\begin{definition}[Scheduling policy]
A \emph{scheduling policy} is an anytime algorithm $f_\mathrm{TS} : V^A \times 2^V \times V^A \to V^A$ that, given the current state and the previous goal map, returns a new goal map within a time budget~$b_{\mathrm{TS}}>0$:
$\rho_{t} = f_\mathrm{TS} \left( x_t, C_t, \rho_{t-1} \,;\, b_{\mathrm{TS}} \right).$
%
%
At $t=0$, $\rho_0$ is initialized by the starting position of agents (i.e., $\rho_0 = x_0$). 
\end{definition}
For agent $a$, $\rho_t(a) = c \in C_t$ indicates an assignment to task $c$, 
while $\rho_t(a) = x_t(a)$ indicates a stay-put command. 
We require that the goal maps be injective so that no two agents are assigned the same goal (in particular, not the same task). 
The goal maps are used as the interface between the scheduling and planning policies.
%
%
\begin{definition}[Planning policy]
A \emph{planning policy} is an anytime algorithm $f_\mathrm{PP}: V^A \times V^A \to V^A$ that takes a system state and a goal map and returns a new joint position:
$x_{t+1} = f_\mathrm{PP} \left( x_t, \rho_t \,;\, b_{\mathrm{PP}} \right),$
%
%
where $b_{\mathrm{PP}} > 0$ is a time budget for planning. 
$f_\mathrm{PP}$ is constrained such that for all $a\in A$, $(x_t(a),x_{t+1}(a)) \in E$, and the transition $(x_t,x_{t+1})$ is collision-free in the standard \gls{acr:mapf} sense (no vertex collisions or swap conflicts).
\end{definition}
\rev{The planner seeks to optimize the agents' paths while guaranteeing deadlock avoidance.}
We refer the reader to Li et al.~\cite{li2025fico} for formal definitions of collisions and conflicts in \gls{acr:mapf}.
The objective in the \gls{acr:lts} problem we study in this article is the maximization of throughput.
\begin{definition}[Throughput]
The \emph{throughput} $\alpha$ is defined as the total number of tasks completed by all agents over the horizon $T_{\max}$. Because the task generator immediately replaces completed tasks, we can write
$\alpha = \sum_{t=0}^{T_{\max}} \sum_{a\in A} \mathbf{1}_{\{ x_t(a) \in C_t \}}.$
%
%
\end{definition}
\paragraph*{Real-Time Execution} 
We target real-time settings (e.g., warehouse fleets) with a fixed per-timestep time budget of $1\,\mathrm{s}$ shared by scheduling and planning policies:
~$b_\mathrm{TS} + b_\mathrm{PP} \leq 1\,\mathrm{s}\, .$
%
%
At each timestep $t$, the scheduler $f_\mathrm{TS}$ computes $\rho_t$ within $b_\mathrm{TS}$ and the planner $f_\mathrm{PP}$ updates $x_{t+1}$. 
Movements are executed and $C_{t+1}$ is updated by the task generator $f_\mathrm{TG}$. 
The problem is initialized by specifying a valid starting position $x_0$ and an initial task set $C_0$.


\subsection{Problem Configuration}
\label{sec:configuration}

The \gls{acr:ts} model above admits a broad family of configurations parametrized by the tuple~$\tup{G, \vert A \vert, M, f_{\mathrm{TG}}, f_{\mathrm{PP}}}$.
In this \rev{work,} we instantiate~$G$ as a warehouse-style layout: a two-dimensional, four-connected grid over traversable \emph{aisle} cells, with shelving modeled by removing blocked cells from~$V$; edges may be directed to encode one-way aisles, and self-loops~$(v,v)\in E$ encode waiting.

To summarize load and spatial density, we use two dimensionless ratios:
~$r_{\mathrm{task/agent}} = M/{\vert A\vert}$,~$r_{\mathrm{agent/node}} = |A|/{\vert V \vert}$.
%
%
The task-to-agent ratio~$r_{\mathrm{task/agent}}$ measures \emph{task pressure} per agent: when~$r_{\mathrm{task/agent}}>1$, a persistent backlog is inevitable, and new tasks cannot all be assigned immediately.
Conversely,~$r_{\mathrm{task/agent}}\leq 1$ allows instantaneous assignment in principle.
The agent-to-node ratio~$r_{\mathrm{agent/node}}$ captures \emph{occupancy}: sparse, lightly loaded regimes (low $r_{\mathrm{agent/node}}$) yield weak coupling and few conflicts; dense regimes (high $r_{\mathrm{agent/node}}$) induce strong coupling, making both scheduling and collision-aware planning critical for throughput.
Directed aisles can mitigate head-on conflicts but increase path asymmetry, further amplifying the role of the planner at higher densities~\cite{zang2025online}.
Our study will explore this design space by sweeping~$(\lvert A\rvert, M)$ \rev{for fixed~$G$, $f_{\mathrm{TG}}$, and $f_{\mathrm{PP}}$, covering sparse-to-dense and under- to over-saturated regimes.}

%% file: sections/06-method.tex
\section{Methodology}
\label{sec:method}

We decompose \gls{acr:ts} into three stages: (i) a data-driven \emph{global guidance} that prescribes a desired distribution of \emph{free} agents across aggregated regions; (ii) a region-to-region \emph{rebalancing} step that solves an optimal transport problem to route agent mass toward the desired distribution; and (iii) \emph{specific task assignments} obtained by solving decoupled local \glspl{acr:ilp} consistent with the rebalancing flow.

We operate on an aggregated graph defined by a seed set~$V_{\mathrm{agg}} \subseteq V$.
Each node~$v\in V$ is mapped to its aggregate region~$\pi(v)\in V_{\mathrm{agg}}$ (e.g., Voronoi partition over seeds, with a fixed tie-breaker).
Region~$i\in V_{\mathrm{agg}}$ thus represents the cell~$\{v\in V\colon \pi(v)=i\}$.


\subsection{Global Guidance}
\label{sec:guidance}

Rather than outputting an assignment directly (i.e., a mapping between free agents and free tasks), the guidance layer produces a low-dimensional intermediate target that shapes \gls{acr:ts}.
Let~$A_t^{f}\subseteq A$ be the set of \emph{free} agents at time~$t$ (i.e., agents eligible to receive a new goal) and let~$N_t\coloneqq \vert A_t^{f}\vert$.
We define the map from the current system state $\tup{x_t, C_t, \rho_{t-1}}$ to a distribution $\delta_t^{d}$ as
$f_{\mathrm{guide}}: V^{A}\times 2^{V}\times V^{A} \to \Delta(V_\mathrm{agg}),$
%
%
where~$\Delta(V_{\mathrm{agg}})=\{\delta\in\mathbb{R}_{+}^{|V_{\mathrm{agg}}|}:\sum_{i}\delta(i)=1\}$ is the probability simplex over regions.
We interpret~$\delta_t^{d}$ as the \emph{desired distribution} of free agents across regions. 
The current free-agent distribution is
\begin{equation}
\delta_t^{f}(i)=\frac{|A_t^{f}\cap \{a:\pi(x_t(a))=i\}|}{N_t}\ \in\ \Delta(V_{\mathrm{agg}}).
\label{eq:free_distribution}
\end{equation}
If~$\delta_t^{d}$ matches the (normalized) distribution of free tasks per region, the method reduces to a greedy assignment; 
when~$\delta_t^{d}$ shifts mass toward regions with few current tasks, it induces proactive rebalancing (e.g., anticipating arrivals or alleviating congestion).


\subsection{Rebalancing}
\label{sec:rebalancing}

Given the desired distribution~$\delta_t^d \in \Delta(V_\mathrm{agg})$ from the global guidance step and the current distribution~$\delta_t^f \in \Delta(V_\mathrm{agg})$ of free agents, the second step computes a flow $y_t$ that transports $\delta_t^f$ to $\delta_t^d$ at minimum cost. 
\rev{We define the integer supplies $n_i^s = N_t \, \delta_t^f(i)$ (cf. \cref{eq:free_distribution}) and the integer demands~$n_i^d = N_t \, \delta_t^d(i)$ (after suitable rounding that ensure $\sum_i n_i^d = N_t$) for all regions~$i \in V_\mathrm{agg}$.}
We model rebalancing as a balanced transportation problem on a \emph{complete} bipartite graph, where left and right node sets are both copies of~$V_\mathrm{agg}$ and the distances~$c_{ij} \in \mathbb{Z}_{\geq 0}$ between nodes $i,j \in V_\mathrm{agg}$ are the shortest-path distances in~$G$. 
Then, the flow~$y_t$ is the solution to the following optimization problem:
\begin{align}
    \min_{y_t} \quad & \sum_{i\in V_\mathrm{agg}}\sum_{j\in V_\mathrm{agg}} c_{ij}\, y_t^{ij} \label{eq:rebalance_obj} \\
    \mathrm{s.t.}\quad 
    & \sum_{j\in V_\mathrm{agg}} y_t^{ij} = n_i^s, && \forall i\in V_\mathrm{agg}, \label{eq:rebalance_supply} \\
    & \sum_{i\in V_\mathrm{agg}} y_t^{ij} = n_j^d, && \forall j\in V_\mathrm{agg}, \label{eq:rebalance_demand} \\
    & y_t^{ij} \in \mathbb{Z}_{\geq 0}, && \forall i,j\in V_\mathrm{agg}. \label{eq:rebalance_integrality}
\end{align}
The decision variables~$y_t^{ij}$ represent the number of agents rebalanced from region~$i$ to region~$j$, and we write~$y_t = (y_t^{ij})_{i,j \in V_\mathrm{agg}}$. 
Since this is a balanced transportation problem with integer supplies, demands, and costs, an optimal integer flow~$y_t$ exists and can be computed efficiently using standard multi-source multi-sink minimum-cost flow solvers~\cite{kovacs2015minimum}.


\subsection{Local Task Assignments}
\label{sec:matching}

Given the region-to-region flow~$y_t$ from the rebalancing step, we compute the goal map~$\rho_t$ by solving $|V_{\mathrm{agg}}|$ \emph{decoupled} local problems, one per region $i\in V_{\mathrm{agg}}$. 
The local problem determines which real agents in region~$i$ (i) satisfy the required outflows to other regions and (ii) are assigned to real tasks in~$i$ consistent with the inflows routed to~$i$.
Let 
$A_i^{f}=\{a\in A :\ a \text{ is free at } t \text{ and } \pi(x_t(a))=i\}$ and $C_i^{f}=\{c\in C_t : \pi(c)=i\}$
%
%
be the \emph{real} free agents and free tasks in region~$i$.
Recall that~$y_t^{ij}$ is the number of free agents to be routed from~$i$ to~$j$.
We denote the total outflow and inflow for region~$i$ by
$\mathrm{out}_i\coloneqq \sum_{j\in V_{\mathrm{agg}}}y_t^{ij}$ and $\mathrm{in}_i\coloneqq \sum_{j\in V_{\mathrm{agg}}}y_t^{ji}$.
%
%
Note that a \emph{self-flow}~$y_t^{ii}$ represents agents that remain in~$i$.

To avoid coupling across regions, cross-region interactions are represented by placeholders located at region seeds:
\begin{itemize}
    \item For each \emph{outgoing} flow $y_t^{ij}$ with $j\neq i$, create an \emph{artificial task set} $C_{j}^{\mathrm{out}}$ with $|C_{j}^{\mathrm{out}}|=y_t^{ij}$, all located at the seed node of region $j$. 
    Intuitively, matching a real agent in $i$ to one of these tasks instructs it to travel to $j$.
    \item For each \emph{incoming} flow $y_t^{ji}$ with $j\neq i$, create an \emph{artificial agent set} $A_{j}^{\mathrm{in}}$ with $|A_{j}^{\mathrm{in}}|=y_t^{ji}$, all located at the seed node of region $j$. Intuitively, matching such a placeholder to a real task in $i$ reserves that task for an agent traveling from $j$.
\end{itemize}

We then aggregate the local sets as 
$A_i = A_i^{f} \cup \big(\bigcup_{j\neq i} A_{j}^{\mathrm{in}}\big)$ and $C_i = C_i^{f} \cup \big(\bigcup_{j\neq i} C_{j}^{\mathrm{out}}\big)$.
%
%
For any~$a\in A_i$ and~$c\in C_i$, let~$w_{ac}\in\mathbb{Z}_{\ge 0}$ denote their directed shortest-path distance in~$G$ (for placeholders we use precomputed seed-to-node distances). 

\paragraph*{Local optimization (region~$i$)}
We formulate a minimum-cost bipartite matching problem:
\begin{align}
    \min_{z^i} \quad & \sum_{a \in A_i} \sum_{c \in C_i} w_{ac}\, z_t^{i,ac} \label{eq:local_obj}\\
    \mathrm{s.t.} \quad
    & \sum_{c \in C_i} z_t^{i,ac} \le 1, && \forall a \in A_i, \label{eq:local_agent}\\
    & \sum_{a \in A_i} z_t^{i,ac} \le 1, && \forall c \in C_i, \label{eq:local_task}\\
    & \sum_{a \in A_i^{f}} z_t^{i,ac} = 1, && \forall c \in \bigcup_{j\neq i} C_{j}^{\mathrm{out}}, \label{eq:local_out_satisfy}\\
    & \sum_{a \in A_i}\sum_{c \in C_i} z_t^{i,ac} = \kappa_i, \label{eq:local_card}\\
    & z_t^{i,ac} \in \{0,1\}, && \forall (a,c)\in A_i\times C_i. \label{eq:local_integrality}
\end{align}
Here,~$z_t^{i,ac}$ indicates whether agent~$a$ is matched to task~$c$, and~$\kappa_i = \min( \vert A_i\vert, \vert C_i\vert )$ is the local matching size.
Constraint \eqref{eq:local_out_satisfy} enforces that every \emph{outgoing} placeholder is fulfilled by a \emph{real} agent currently in~$i$;
equivalently, artificial agents cannot be matched to artificial tasks, preventing spurious cross-region matches.
The remaining constraints are standard assignment constraints.
Because the constraint matrix is totally unimodular, the LP relaxation already yields integral solutions; the problem can be solved as a min-cost flow or with the Hungarian algorithm after adding dummies to make it square~\cite{bertsekas1998network}.

\begin{remark}[Feasibility and interpretation]
Since the rebalancing step conserves free agents, we have~$\sum_i \mathrm{out}_i = \sum_i \mathrm{in}_i = N_t$, and \rev{for each~$i$,} the supply constraint guarantees~$\mathrm{out}_i = |A_i^{f}|$ (recall supplies are derived from actual free-agent counts). 
Thus, the number of outgoing placeholders in~$i$ equals~$|A_i^{f}|$, and \eqref{eq:local_out_satisfy} is feasible. 
Self-flow~$y_t^{ii}$ is handled implicitly: those agents are matched to \emph{real} tasks in~$C_i^{f}$ or are unmatched (i.e., they remain within~$i$ without a task).
\end{remark}

\paragraph*{Recovering the global goal map}
After solving \eqref{eq:local_obj}–\eqref{eq:local_integrality} for all~$i\in V_{\mathrm{agg}}$:
\begin{enumerate}
    \item If a \emph{real agent}~$a\in A_i^{f}$ is matched to a \emph{real task}~$c\in C_i^{f}$, set~$\rho_t(a)=c$.
    \item If a \emph{real agent}~$a_i$ in region $i$ is matched to an outgoing placeholder corresponding to flow from $i$ to $j$, and an incoming placeholder corresponding to flow from $i$ to $j$ is matched to a \emph{real task}~$c_j$ in region $j$ (cf. \cref{fig:overview_diagram}), we set $\rho_t(a_i) = c_j$. When multiple agents traverse the same pair~$(i,j)$, \rev{we pair the list of $a_i$'s with the list of $c_j$'s using a distance-based greedy procedure that iteratively matches each agent to the nearest currently unassigned reserved task in $j$.}
\end{enumerate}
Any real agent that remains unmatched to a real task receives a waypoint at its designated destination region, ensuring a well-defined goal for every free agent.


\section{Method Details}
\label{sec:method_details}

In this section, we spell out the components used by our scheduler: (i) a graph aggregation that partitions the workspace into a small number of regions while preserving shortest-path structure; 
(ii) a data-driven guidance map~$f_{\mathrm{guide}}$ learned with \gls{acr:rl} that outputs a desired distribution~$\delta_t^d$ over regions;
and (iii) the rebalancing and local assignment procedures described in \cref{sec:rebalancing} and \cref{sec:matching}.
Here, we focus on how we construct~$V_{\mathrm{agg}}$, define the state and action spaces for learning, and implement the guidance policy.


\subsection{Graph Aggregation}

Given a graph~$G=\tup{V,E}$, \rev{the aggregation step chooses \emph{seed} nodes~$V_\mathrm{agg} \subseteq V$ and assigns every node~$v\in V$ to one region via a shortest-path Voronoi partition.}
Let~$\operatorname{dist}_G(u,w)$ denote the (directed) shortest-path distance in~$G$ (finite on our warehouse layouts).
We define the regional map
%
\begin{equation*}
    \pi(v) \in \underset{i\in V_{\mathrm{agg}}}{\arg\min} \operatorname{dist}_G(v,i),
\end{equation*}
breaking ties by a fixed, deterministic rule (e.g., lowest seed index), so that the regions~$\{V_i\coloneqq \{v\in V\colon \pi(v)=i\}\}_{i\in V_{\mathrm{agg}}}$ form a partition of~$V$.
 
On dense warehouse layouts, we populate~$V_{\mathrm{agg}}$ with: (i) all aisle intersections (i.e., to capture routing hubs/bottlenecks), and (ii) perimeter pick/pack stations (i.e., to capture sinks/sources).
This yields regions aligned with traffic structure while preserving geodesic distances in~$G$ (including directed one-way aisles and waiting self-loops).
In typical instances, this aggregation reduces the number of vertices (i.e., from $|V|$ to $|V_\mathrm{agg}|$) by more than 90\%.
We show an example of this aggregation in \cref{fig:aggregation}. \rev{More generally, seeds can be chosen by overlaying a coarse, approximately uniform grid on the map.}

\input{figures/plot_aggregation}


\subsection{Global Guidance via Reinforcement Learning}
\label{sec:rl}

We learn a guidance map~$f_\mathrm{guide}$ that outputs a desired regional distribution~$\delta_t^d \in \Delta(V_\mathrm{agg})$. 
A data-driven approach via \gls{acr:gnn} \gls{acr:rl} lets us adapt to arbitrary graphs~$G$, planning policies~$f_{\mathrm{PP}}$, task generators~$f_{\mathrm{TG}}$, and performance metrics.

\paragraph{State Representation} 
We build a sparse neighborhood graph~$G_\mathrm{nh} = \tup{V_\mathrm{agg}, E_\mathrm{nh}}$ over the aggregate regions.
An ordered pair~$\tup{i,j}\in E_{\mathrm{nh}}$ is included when the directed geodesic distance in~$G$ is below a threshold~$\varepsilon >0$, i.e.,~$\tup{i,j}\in E_{\mathrm{nh}} \Longleftrightarrow \operatorname{dist}_G(i,j)\leq \varepsilon,$
%
%
so~$G_{\mathrm{nh}}$ inherits directionality from~$G$ (one-way aisles remain asymmetric).
Given the system state~$\tup{x_t,C_t}$ and the previous goal map~$\rho_{t-1}$, a feature map
\begin{equation*}
    f_{\mathrm{feature}}:\ V^{A}\times 2^{V}\times V^{A} \to 
(\mathbb{R}^{m_v})^{V_{\mathrm{agg}}} \times (\mathbb{R}^{m_e})^{E_{\mathrm{nh}}},
\end{equation*}
produces per-node and per-edge features that serve as the \gls{acr:rl} state~$s_t\in \mathcal{S}$.
Node features summarize local load, capacity, and context: we include counts of agents, free agents, tasks, and free tasks currently in region~$i$, normalized by~$\vert V_i\vert$ or by the number of free agents~$N_t$ for scale invariance.
Further, we include a congestion proxy given by the fraction of unoccupied nodes in~$V_i$ (clipped to $[0,1]$), a short-horizon estimate of incoming and outgoing flow based on the previous goal map, and sinusoidal encodings of the region's spatial position together with a time encoding to disambiguate symmetries and capture periodic demand.
Edge features capture travel and cross-region interaction: we use the directed geodesic length~$\ell_{ij}=\operatorname{dist}_G(i,j)$ together with its bounded reciprocal~$1/(1+\ell_{ij})$.
Further, we consider a demand–supply hint equal to the number of free tasks in $j$ that are closer (in $G$) to free agents in $i$ than any other agents, normalized by $\vert V_j\vert$, and a corridor-load statistic computed along a representative shortest path from $i$ to $j$ (e.g., mean or maximum occupancy, again clipped to $[0,1]$).

\paragraph{Algorithm and Policy Representation} 
We train the guidance policy with \gls{acr:sac}~\cite{haarnoja2018soft}, a model-free, off-policy \gls{acr:rl} algorithm with entropy regularization. 
The (parameterized) actor~$f_\mathrm{actor}^\theta : \mathcal{S} \to \mathcal{A}$ maps the \gls{acr:rl} state to the \gls{acr:rl} action space~$\mathcal{A} \coloneqq \mathbb{R}_+^{|V_\mathrm{agg}|}$. 
For a state $s_t \in \mathcal{S}$, we interpret the output of the actor $f_\mathrm{actor}^\theta(s_t) = a_t$ as a concentration parameter for a Dirichlet distribution $\mathrm{Dir}(a_t)$, where each sample $\alpha \sim \mathrm{Dir}(a_t)$ is a valid probability vector (i.e., $\alpha \in \Delta(V_\mathrm{agg})$). During training, \gls{acr:sac} keeps the actor stochastic, while at deployment, we operate based on $\mathbb{E}_{\alpha \sim \mathrm{Dir}(a_t)}[\alpha]$. 
Because~$s_t$ is defined on the neighborhood graph~$G_\mathrm{nh}$, both actor and critic share a graph-aware encoder: a transformer-style graph network (i.e., attention with neighborhood masking~\cite{shi2021masked}) followed by per-node MLPs.
The critic includes an additional permutation-invariant aggregation (sum/mean pooling) to obtain a global embedding before the value head.
The critic~$f_\mathrm{critic}^\theta : \mathcal{S} \times \mathcal{A} \to \mathbb{R}$ thus consumes~$\tup{s_t, a_t}$ and estimates~$Q(s_t,a_t)$.

\paragraph{Reward Design} 
The guidance policy is trained with a shaped reward that balances immediate completions and proximity to completion.
Let~$D_t=\{c\in C_t\colon \exists a\in A \text{ s.t. }x_t(a)=c\}$ be the set of tasks finished at time~$t$ (cf. throughput in \cref{sec:math_model}).
The completion term is~$r_t^\mathrm{fin}=\vert D_t\vert$ (optionally normalized by~$\vert A\vert$ for scale invariance).
To mitigate sparsity-pronounced at high~$r_{\mathrm{task/agent}}$ or~$r_{\mathrm{agent/node}}$, we add a progress term that rewards states in which assigned tasks are close to finishing. Let~$A_t^\mathrm{act} = \{ a \in A \colon \rho_t(a) \in C_t \}$ denote the set of agents actively assigned to a task at time $t$ and $\tau_t(a) \coloneqq \min_{\tau} \{ \tau \geq t : x_\tau(a) = \rho_t(a) \}$ denote the first completion time. We set $r_t^\mathrm{fut} = \sum_{ a \in A_t^\mathrm{act}} \phi(\tau_t(a)-t)$ with~$\phi(t)=1/(1+t)$ or~$\phi(t)=e^{-t/\kappa}$, so~$r_t^\mathrm{fut}$ is large when many active assignments are near completion. We delay enqueuing the transition~$\tup{s_t,a_t,r_t,s_{t+1}}$ into the replay buffer until~$r_t^\mathrm{fut}$ is available (i.e., after observing a long enough trace of $x_t$).
The final reward is a weighted sum~$r_t = c_1 r_t^{\mathrm{fin}} + c_2 r_t^{\mathrm{fut}}$, with~$c_1,c_2>0$ chosen by a grid search on a validation set.


\subsection{Optimization Solver}
Both the global rebalancing in \cref{sec:rebalancing} and the local assignments in \cref{sec:matching} are implemented with off-the-shelf minimum-cost flow / bipartite matching solvers on their \emph{linear} relaxations. 
\paragraph{\rev{Global Rebalancing}}
We pose rebalancing as a balanced transportation problem on the complete graph over~$V_{\mathrm{agg}}$, with $\vert V_{\mathrm{agg}}\vert$ supply/demand nodes and $\vert V_{\mathrm{agg}}\vert^2$ arcs. 
Standard min-cost flow algorithms solve this in~$O(\vert V_{\mathrm{agg}}\vert^3)$ time in the worst case.
\paragraph{\rev{Local Assignment}}
The matching stage decomposes across regions~$i \in V_{\mathrm{agg}}$. 
Let~$n_i \coloneqq \max\{\vert A_i\vert,\vert C_i\vert\}$. 
After padding to an~$n_i \times n_i$ cost matrix, each regional problem is solvable in~$O(n_i^3)$ time (e.g., Hungarian) or as a small min-cost flow instance with identical complexity guarantees. 
These subproblems are \emph{fully parallelizable} across regions; with sufficient workers, the wall-clock time is dominated by the largest region:
\begin{equation*}
\text{serial: } O\!\Big(\vert V_{\mathrm{agg}}\vert^3 + \sum_{i} n_i^3\Big),
\quad
\text{parallel: } O\!\Big(\vert V_{\mathrm{agg}}\vert^3 + \max_{i} n_i^3\Big).
\end{equation*}

%% file: figures/plot_aggregation.tex
\begin{figure}[t]
\setlength{\abovecaptionskip}{0pt}
\centering
\includegraphics[width=0.45\columnwidth]{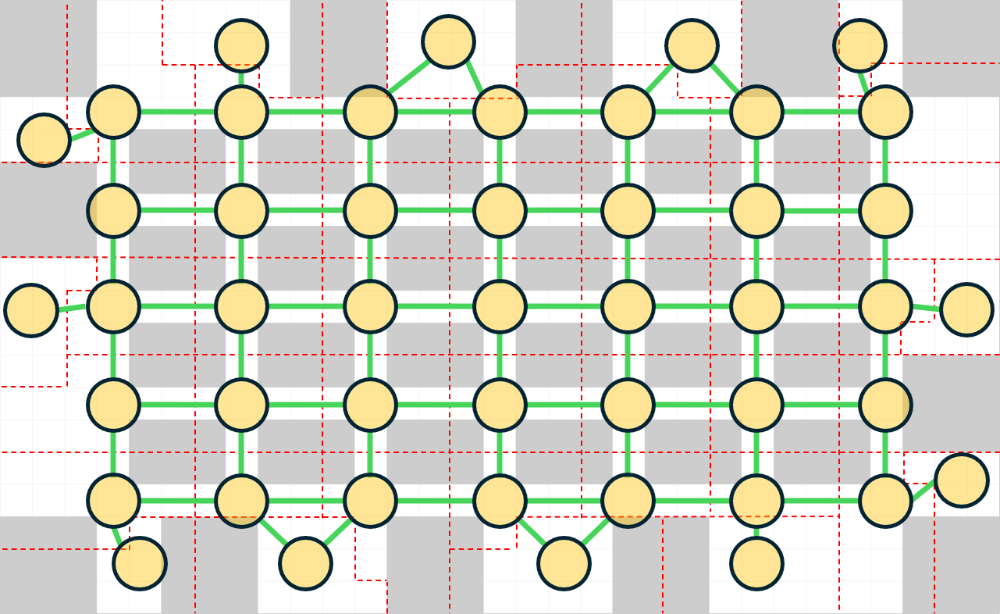} 
\caption{An example of our graph aggregation for warehouse layouts.}
\label{fig:aggregation}
\end{figure}

%% file: sections/07-results.tex
\section{Case Study}

To validate our method and compare it to the state-of-the-art, we perform a case study on different warehouse scenarios in simulation.


\subsection{Environment}

We use \gls{acr:lrr}~\cite{chan2024league} as the environment for all numerical results.
\gls{acr:lrr} \rev{provides a simulation engine for} \gls{acr:mapf} and \gls{acr:ts}, enabling fair comparisons across algorithms. 
The simulator in \gls{acr:lrr} assumes a discrete-time, 4-neighbor grid with heading-constrained motion, where at each step an agent may rotate in place, move forward one cell, or wait. 
Tasks in \gls{acr:lrr} \rev{require an agent} to visit two locations (errands) in order (i.e., \gls{acr:mapd}). 
Agents may be \emph{reassigned} to a different task provided they have not yet started the first errand.

The \gls{acr:lrr} effort also includes an annual competition in which teams submit new algorithms for \gls{acr:mapf} and \gls{acr:ts}.
For the planning policy~$f_\mathrm{PP}$, we use the \gls{acr:tfgp} algorithm~\cite{chen2024traffic}, the default planner in the 2024 \gls{acr:lrr} competition. 
For the task generator~$f_\mathrm{TG}$, we uniformly sample new tasks from the set of unoccupied nodes at time~$t$. 
\rev{Throughout, we report the number of tiles~$\vert V_\mathrm{tile}\vert$ as the size of the graph $G$. A tile corresponds to a node in the discretized environment and may be either traversable (e.g., warehouse pathways) or non-traversable (e.g., shelves).}
Unless stated otherwise, $T_{\max}=10,000$.
The remaining parameters~$\tup{\vert V_\mathrm{tile}\vert, \vert A \vert, M}$ vary across figures.
\rev{
\begin{remark}[Success rate]
In all tested scenarios, we ensure sufficient time for the planner~$f_{\mathrm{PP}}$ to return a valid plan.
As a result, we do not evaluate regimes of extreme congestion that may induce deadlocks, which would primarily reflect properties of the planner rather than the scheduler.
\end{remark}
}


\subsection{Benchmarks}

The 2024 \gls{acr:lrr} competition concluded in March 2025, with 22 teams participating in the \gls{acr:ts} track. 
We compare against three baselines.
First, \ALGLRR, the winning \gls{acr:ts} entry by team \emph{No Man's Sky}~\cite{yukhnevich2025enhancing}, is a heuristic that ranks agent-task pairs using a customized cost metric.
Second, \ALGILP, a \emph{global} minimum-cost matching between all free agents and available tasks, where edge costs are shortest-path distances.
Although the problem is posed as an \gls{acr:ilp}, we solve its linear relaxation, which is tight for assignment and yields an integral optimum (equivalently solvable by the Hungarian method).
Finally, \ALGG, the default greedy assignment implemented in \gls{acr:lrr}~\cite{chan2024league}.
We refer to these methods as \ALGLRR, \ALGILP, and \ALGG.


\subsection{Training and Compute}

We train the \gls{acr:rl} guidance map~$f_\mathrm{guide}$ with a horizon~$T_{\max}=150$.
Training converges in roughly 2,000 episodes (around 24 hours wall-clock) on a single NVIDIA Quadro RTX 8000 (48 GB VRAM).
With our choice of $\varepsilon$, the \gls{acr:rl} graph is highly connected, containing about 90\% of the edges of the corresponding complete graph.
Empirically, using a shorter horizon with more episodes stabilizes learning and improves final performance.
For evaluation, \emph{all methods} are executed under an identical compute budget: an AMD Ryzen 7 5825U CPU (single-threaded) with 16 GB RAM.
To ensure fairness, we neither parallelize the local assignment step nor run~$f_{\mathrm{guide}}$ on a GPU during evaluation.


\subsection{Throughput}

\input{figures/plot_throughput}

\cref{fig:throughput} reports throughput as a function of the number of agents~$|A|$ and map size~$|V_\mathrm{tile}|$.
Across all configurations we fix the environment parameters to~$r_{\mathrm{task/agent}} = 1.5$ and~$r_{\mathrm{agent/node}} = 0.35$.
For our method, we train a separate policy for each~$\tup{\vert A \vert, \vert V_{\mathrm{tile}}}$ using a single set of hyperparameters.
Our algorithm outperforms all benchmarks at every scale: relative to \ALGLRR, the average gains are about 10\% on the two medium instances and 4.3\% on the largest instance. When we consider a more restrictive setting in which \gls{acr:ts} algorithms are not permitted to reassign tasks, the performance gap between our algorithm and competing methods widens, as shown in~\cref{table:task_change}.

\input{figures/table_task_change}
\input{figures/table_time}
\input{figures/plot_heat_map}

To probe the source of these gains, \cref{table:time} decomposes the average \emph{time-to-task} (i.e., time to reach the first errand) and \emph{time-in-task} (i.e., time from the first to the second errand) for~$\vert A \vert=200$ and~$\vert V_{\mathrm{tile}}\vert=975$.
The results indicate that higher throughput does not arise from merely dispatching the nearest free agents.
Instead, agents complete assigned tasks faster under our \gls{acr:ts} policy, pointing to reduced congestion.
We quantify congestion by counting how often the planner~$f_{\mathrm{PP}}$ must deviate from its preferred path and execute a lower-priority action (i.e., a \emph{conflict}).
\cref{fig:conflict} visualizes this conflict density for our method and for \ALGLRR.
On this instance, our method reduces the peak number of conflicts by 23\% and the total number of conflicts by 20\%, consistent with the observed throughput improvements. \rev{Finally, in \cref{table:maps}, we report throughput across a range of map topologies beyond warehouses \frev{(shown in \cref{fig:other_maps})}. The results show that our method remains competitive across diverse environments.}

\input{figures/table_maps}
\input{figures/other_maps}


\subsection{Scheduling Time}

\input{figures/plot_cpu_time}

In \cref{fig:cpu_time}, we compare the \gls{acr:ts} per-step computation times across different scales. 
Specifically, we distinguish between the initial steps in the environment, where all agents are unassigned, and the lifelong (steady state) steps, where the population of free agents and open tasks stabilizes. 
As expected, optimization-based baselines incur higher latency than heuristic methods.
In steady state, our approach dedicates over 90\% of the $1\,\mathrm{s}$ control-cycle budget to the planner~$f_{\mathrm{PP}}$ and still runs 
much
faster than the global matching baseline \ALGILP.
For large deployments, wall-clock can be further reduced through engineering, via accelerators for the guidance module (e.g., TPU/ASIC) and parallelization of the local assignment step, while keeping the evaluation setting 
unchanged.


\subsection{Adaptability}

\input{figures/plot_generalize_agent}
\input{figures/plot_generalize_size}

Because our approach involves learning, we evaluate \emph{zero-shot} transfer: train on a single configuration and test on different instances \emph{without} retraining or tuning.
\cref{fig:generalize_agent} probes sensitivity to the ratio~$r_{\mathrm{agent/node}}$, \cref{fig:generalize_size} probes sensitivity to graph size~$\vert V_{\mathrm{tile}}\vert$.
Across both stressors, our \gls{acr:ts} policy adapts to unseen settings and, in most cases, continues to outperform the state-of-the-art \ALGLRR baseline. 
Notably, performance is competitive even when the evaluation conditions depart substantially from the training configuration, indicating that the learned guidance captures transferable structure rather than overfitting to a single regime.


\subsection{\rev{Ablation}}

\input{figures/table_ablation}

\rev{
In \cref{table:ablation}, we present an ablation study of the key components of our method.
For the guidance step, we replace the \gls{acr:rl} distribution with (i) a uniform distribution over nodes, (ii) a demand distribution formed by normalizing the number of free tasks at each node, or (iii) a random Dirichlet distribution with parameter $\alpha=0.1$. The \gls{acr:rl} distribution outperforms all three alternatives.
Notably, the uniform distribution yields throughput comparable to the global optimization baseline. Intuitively, uniform guidance imposes no directional bias on agent movements, reducing our framework to a linear assignment problem.
Even with weakened guidance, the system remains well-behaved, since feasibility and collision avoidance are enforced downstream rather than in the learned component.
For the matching step, we replace the decoupled linear assignment problems with a flow-constrained greedy matching. While this improves over the unguided greedy baseline \ALGG (cf. \cref{table:task_change}), it incurs a substantial drop in throughput relative to our matching procedure 
in \cref{sec:matching}. 
}

%% file: figures/plot_throughput.tex
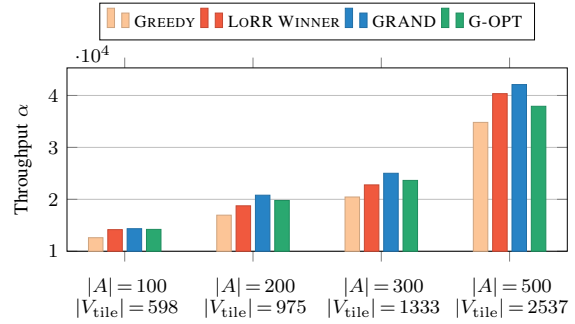
\begin{figure}[t]
\setlength{\abovecaptionskip}{-2pt}
\begin{adjustbox}{max width=\Scale\columnwidth, center}
\begin{tikzpicture}
\begin{axis}[
  ybar,
  bar width=6pt,
  width=\columnwidth, height=4.25cm, 
  ymin=10000,
  ylabel={Throughput~$\alpha$},
  ylabel style={font=\footnotesize},
  yticklabel style={font=\footnotesize},
  symbolic x coords={
    {$|A|\,{=}\,100$\\$|V_\mathrm{tile}|\,{=}\,598 $},
    {$|A|\,{=}\,200$\\$|V_\mathrm{tile}|\,{=}\,975 $},
    {$|A|\,{=}\,300$\\$|V_\mathrm{tile}|\,{=}\,1333$},
    {$|A|\,{=}\,500$\\$|V_\mathrm{tile}|\,{=}\,2537$}
  },
  xtick=data,
  xticklabel style={align=center, text width=2.5cm, anchor=east, yshift=-15pt, xshift=40pt, font=\footnotesize},
  legend style={at={(0.5,1.15)}, anchor=south, legend columns=-1, font=\scriptsize},
  enlarge x limits=0.15,
  ymajorgrids=true,
]
\addplot+[styleGreedy] coordinates {
  ({$|A|\,{=}\,100$\\$|V_\mathrm{tile}|\,{=}\,598 $}, 12610)
  ({$|A|\,{=}\,200$\\$|V_\mathrm{tile}|\,{=}\,975 $}, 16952)
  ({$|A|\,{=}\,300$\\$|V_\mathrm{tile}|\,{=}\,1333$}, 20449)
  ({$|A|\,{=}\,500$\\$|V_\mathrm{tile}|\,{=}\,2537$}, 34816)
};
\addplot+[styleLRR] coordinates {
  ({$|A|\,{=}\,100$\\$|V_\mathrm{tile}|\,{=}\,598 $}, 14164)
  ({$|A|\,{=}\,200$\\$|V_\mathrm{tile}|\,{=}\,975 $}, 18752)
  ({$|A|\,{=}\,300$\\$|V_\mathrm{tile}|\,{=}\,1333$}, 22781)
  ({$|A|\,{=}\,500$\\$|V_\mathrm{tile}|\,{=}\,2537$}, 40350)
};
\addplot+[styleRL] coordinates {
  ({$|A|\,{=}\,100$\\$|V_\mathrm{tile}|\,{=}\,598 $}, 14344)
  ({$|A|\,{=}\,200$\\$|V_\mathrm{tile}|\,{=}\,975 $}, 20796)
  ({$|A|\,{=}\,300$\\$|V_\mathrm{tile}|\,{=}\,1333$}, 25018)
  ({$|A|\,{=}\,500$\\$|V_\mathrm{tile}|\,{=}\,2537$}, 42098)
};
\addplot+[styleILP] coordinates {
  ({$|A|\,{=}\,100$\\$|V_\mathrm{tile}|\,{=}\,598 $}, 14211)
  ({$|A|\,{=}\,200$\\$|V_\mathrm{tile}|\,{=}\,975 $}, 19781)
  ({$|A|\,{=}\,300$\\$|V_\mathrm{tile}|\,{=}\,1333$}, 23644)
  ({$|A|\,{=}\,500$\\$|V_\mathrm{tile}|\,{=}\,2537$}, 37918)
};
\legend{\ALGG, \ALGLRR, \ALGRL, \ALGILP}
\end{axis}
\end{tikzpicture}
\end{adjustbox}
\caption{Throughput for varying numbers of agents~$|A|$ and map sizes~$|V_\mathrm{tile}|$.}
\label{fig:throughput}
\end{figure}

%% file: figures/table_task_change.tex
\begin{table}
\setlength{\abovecaptionskip}{0pt}
\centering
\footnotesize
\renewcommand{\arraystretch}{\TableVerStretch} 
\caption{Throughput for 200 agents and a map size of 975.}
\begin{adjustbox}{max width=\Scale\columnwidth, center}
\begin{NiceTabular}{c|c|c}
    \toprule
    \textbf{Method} & Without Reassignment & With Reassignment \\
    \midrule
    \ALGG   & 12372 & 16952 \\
    \ALGLRR & 12990 & 18752 \\
    \ALGRL  & \textbf{17916} & \textbf{20752} \\
    \ALGILP & 16554 & 19781 \\
    \bottomrule
\end{NiceTabular}
\end{adjustbox}
\label{table:task_change}
\end{table}

%% file: figures/table_time.tex
\begin{table}[t]
\setlength{\abovecaptionskip}{0pt}
\centering
\footnotesize
\renewcommand{\arraystretch}{\TableVerStretch} 
\caption{Throughput, time-to-task, and time-in-task for 200 agents.}
\begin{adjustbox}{max width=\Scale\columnwidth, center}
\begin{NiceTabular}{c|c|c|c|c}
    \toprule
    \textbf{Metric} & \ALGG & \ALGLRR & \ALGRL  & \ALGILP  \\
    \midrule
    \textbf{Throughput} & 16952 & 18752 & \textbf{20752} & 19781 \\
    \textbf{Time-to-task} & 32.6 s & 16.1 s & 24.0 s & \textbf{10.2 s} \\
    \textbf{Time-in-task} & 86.9 s & 88.5 s & \textbf{70.9 s} & 90.6 s \\
    \bottomrule
\end{NiceTabular}
\end{adjustbox}
\label{table:time}
\end{table}

%% file: figures/plot_heat_map.tex
\begin{figure}
\setlength{\abovecaptionskip}{0pt}
\centering
\hspace{12pt}
\begin{minipage}[b]{0.33\columnwidth}
\centering
\includegraphics[width=\columnwidth]{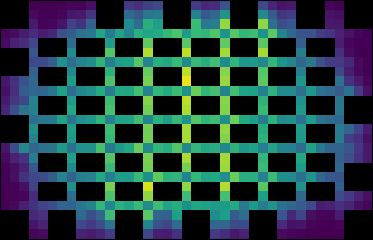}
{\footnotesize (a) \ALGLRR}
\end{minipage}
\hfill
\begin{minipage}[b]{0.33\columnwidth}
\centering
\includegraphics[width=\columnwidth]{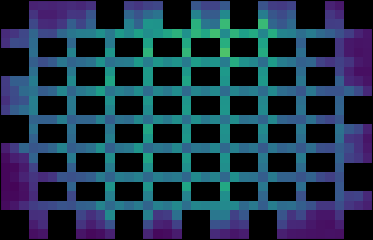}
{\footnotesize (a) \ALGRL}
\end{minipage}
\hspace{2pt}
\begin{minipage}[b]{0.0635\columnwidth}
\centering
\includegraphics[width=\columnwidth]{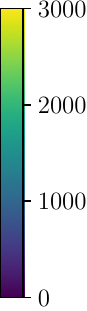}
\vspace{-2.5pt}
\end{minipage}
\hspace{12pt}
\caption{\rev{Number of total conflicts for $(|A|, |V_\mathrm{tile}|) = (200, 975)$.}}
\label{fig:conflict}
\end{figure}

%% file: figures/table_maps.tex
\begin{table}[t]
\setlength{\tabcolsep}{3.65pt}
\setlength{\abovecaptionskip}{0pt}
\centering
\footnotesize
\renewcommand{\arraystretch}{\TableVerStretch} 
\rev{
\caption{Throughput across different map topologies.}
\begin{adjustbox}{max width=\Scale\columnwidth, center}
\begin{NiceTabular}{c|c|c|c|c}
    \toprule
    \textbf{Map} & \ALGG & \ALGLRR & \ALGRL  & \ALGILP  \\
    \midrule
    \Block{1-1}{\textbf{Room} $\boldsymbol{(|A|\,{=}\,200)}$} & 10993 & 12561 & \textbf{14912} & 13801 \\
    \midrule
    \Block{1-1}{\textbf{Maze} $\boldsymbol{(|A|\,{=}\,400})$} & 1508 & 2013 & \textbf{2417} & 2062 \\
    \bottomrule
\end{NiceTabular}
\end{adjustbox}
\label{table:maps}
}
\end{table}

%% file: figures/other_maps.tex
\begin{figure}
\setlength{\abovecaptionskip}{0pt}
\centering
\begin{minipage}[b]{0.25\columnwidth}
\centering
\includegraphics[width=\columnwidth]{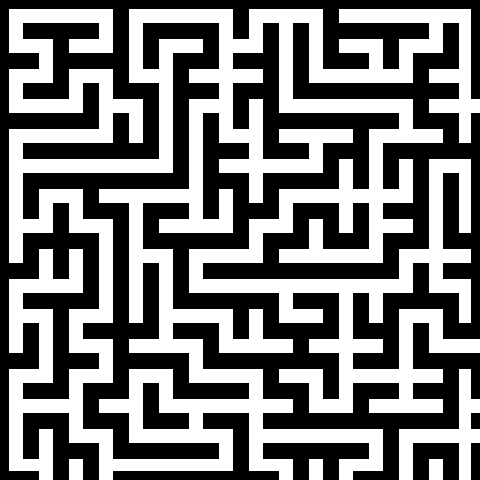}
\end{minipage}
\hspace{24pt}
\begin{minipage}[b]{0.25\columnwidth}
\centering
\includegraphics[width=\columnwidth]{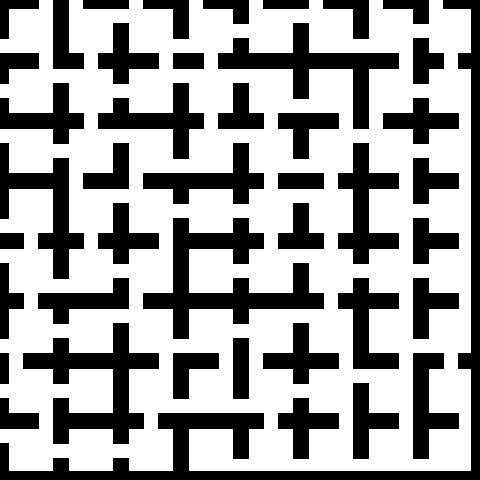}
\end{minipage}
\caption{\frev{Visualization of the Maze (left) and Room (right) environments.}}
\label{fig:other_maps}
\end{figure}

%% file: figures/plot_cpu_time.tex
\begin{figure}[t!]
\setlength{\abovecaptionskip}{-2pt}
\begin{adjustbox}{max width=\Scale\columnwidth, center}
\begin{tikzpicture}
\begin{groupplot}[
  group style={
    group size=1 by 2,
    vertical sep=1.5cm,
  },
  ybar,
  stack plots=y,
  /pgf/bar width=6pt,
  width=\columnwidth, height=3.25cm, 
  ymin=0,
  ylabel style={font=\footnotesize},
  yticklabel style={font=\footnotesize},
  legend style={font=\scriptsize}, 
  symbolic x coords={
    {$|A|\,{=}\,100$\\$|V_\mathrm{tile}|\,{=}\,598 $},
    {$|A|\,{=}\,200$\\$|V_\mathrm{tile}|\,{=}\,975 $},
    {$|A|\,{=}\,300$\\$|V_\mathrm{tile}|\,{=}\,1333$},
    {$|A|\,{=}\,500$\\$|V_\mathrm{tile}|\,{=}\,2537$}
  },
  xtick=data,
  xticklabel style={align=center, text width=3cm, font=\footnotesize},
  legend style={at={(0.5,2.54)}, anchor=south, legend columns=4, font=\tiny},
  enlarge x limits=0.2,
  ymajorgrids=true,
]

\nextgroupplot[
  ylabel={\rev{Initial (ms)}},
  ymax=1100,
]
\addplot+[styleGreedy, bar shift=-12pt] coordinates {
  ({$|A|\,{=}\,100$\\$|V_\mathrm{tile}|\,{=}\,598 $}, 1)
  ({$|A|\,{=}\,200$\\$|V_\mathrm{tile}|\,{=}\,975 $}, 2)
  ({$|A|\,{=}\,300$\\$|V_\mathrm{tile}|\,{=}\,1333$}, 8)
  ({$|A|\,{=}\,500$\\$|V_\mathrm{tile}|\,{=}\,2537$}, 33)
};
\resetstackedplots
\addplot+[styleLRR, bar shift=-4pt] coordinates {
  ({$|A|\,{=}\,100$\\$|V_\mathrm{tile}|\,{=}\,598 $}, 14)
  ({$|A|\,{=}\,200$\\$|V_\mathrm{tile}|\,{=}\,975 $}, 20)
  ({$|A|\,{=}\,300$\\$|V_\mathrm{tile}|\,{=}\,1333$}, 30)
  ({$|A|\,{=}\,500$\\$|V_\mathrm{tile}|\,{=}\,2537$}, 61)
};
\resetstackedplots
\addplot+[styleRL, bar shift=+4pt] coordinates {
  ({$|A|\,{=}\,100$\\$|V_\mathrm{tile}|\,{=}\,598 $}, 11)
  ({$|A|\,{=}\,200$\\$|V_\mathrm{tile}|\,{=}\,975 $}, 15)
  ({$|A|\,{=}\,300$\\$|V_\mathrm{tile}|\,{=}\,1333$}, 22)
  ({$|A|\,{=}\,500$\\$|V_\mathrm{tile}|\,{=}\,2537$}, 26) 
};
\addplot+[styleRL, bar shift=+4pt, forget plot, fill opacity=0.66] coordinates {
  ({$|A|\,{=}\,100$\\$|V_\mathrm{tile}|\,{=}\,598 $}, 12)
  ({$|A|\,{=}\,200$\\$|V_\mathrm{tile}|\,{=}\,975 $}, 14)
  ({$|A|\,{=}\,300$\\$|V_\mathrm{tile}|\,{=}\,1333$}, 21)
  ({$|A|\,{=}\,500$\\$|V_\mathrm{tile}|\,{=}\,2537$}, 58) 
};
\addplot+[styleRL, bar shift=+4pt, forget plot, fill opacity=0.33] coordinates {
  ({$|A|\,{=}\,100$\\$|V_\mathrm{tile}|\,{=}\,598 $}, 55)
  ({$|A|\,{=}\,200$\\$|V_\mathrm{tile}|\,{=}\,975 $}, 77)
  ({$|A|\,{=}\,300$\\$|V_\mathrm{tile}|\,{=}\,1333$}, 110)
  ({$|A|\,{=}\,500$\\$|V_\mathrm{tile}|\,{=}\,2537$}, 326) 
};
\resetstackedplots
\addplot+[styleILP, bar shift=+12pt]  coordinates {
  ({$|A|\,{=}\,100$\\$|V_\mathrm{tile}|\,{=}\,598 $}, 39)
  ({$|A|\,{=}\,200$\\$|V_\mathrm{tile}|\,{=}\,975 $}, 187)
  ({$|A|\,{=}\,300$\\$|V_\mathrm{tile}|\,{=}\,1333$}, 372)
  ({$|A|\,{=}\,500$\\$|V_\mathrm{tile}|\,{=}\,2537$}, 1073)
};

\nextgroupplot[
  ylabel={\rev{Lifelong (ms)}},
  ymax=330,
]
\addplot+[styleGreedy, bar shift=-12pt] coordinates {
  ({$|A|\,{=}\,100$\\$|V_\mathrm{tile}|\,{=}\,598 $}, 1)
  ({$|A|\,{=}\,200$\\$|V_\mathrm{tile}|\,{=}\,975 $}, 1)
  ({$|A|\,{=}\,300$\\$|V_\mathrm{tile}|\,{=}\,1333$}, 1)
  ({$|A|\,{=}\,500$\\$|V_\mathrm{tile}|\,{=}\,2537$}, 1)
};
\resetstackedplots
\addplot+[styleLRR, bar shift=-4pt] coordinates {
  ({$|A|\,{=}\,100$\\$|V_\mathrm{tile}|\,{=}\,598 $}, 2)
  ({$|A|\,{=}\,200$\\$|V_\mathrm{tile}|\,{=}\,975 $}, 3)
  ({$|A|\,{=}\,300$\\$|V_\mathrm{tile}|\,{=}\,1333$}, 3)
  ({$|A|\,{=}\,500$\\$|V_\mathrm{tile}|\,{=}\,2537$}, 3)
};
\resetstackedplots
\addplot+[styleRL, bar shift=+4pt] coordinates {
  ({$|A|\,{=}\,100$\\$|V_\mathrm{tile}|\,{=}\,598 $}, 5)
  ({$|A|\,{=}\,200$\\$|V_\mathrm{tile}|\,{=}\,975 $}, 8)
  ({$|A|\,{=}\,300$\\$|V_\mathrm{tile}|\,{=}\,1333$}, 8)
  ({$|A|\,{=}\,500$\\$|V_\mathrm{tile}|\,{=}\,2537$}, 17)
};
\addplot+[styleRL, bar shift=+4pt, forget plot, fill opacity=0.66] coordinates {
  ({$|A|\,{=}\,100$\\$|V_\mathrm{tile}|\,{=}\,598 $}, 4)
  ({$|A|\,{=}\,200$\\$|V_\mathrm{tile}|\,{=}\,975 $}, 5)
  ({$|A|\,{=}\,300$\\$|V_\mathrm{tile}|\,{=}\,1333$}, 7)
  ({$|A|\,{=}\,500$\\$|V_\mathrm{tile}|\,{=}\,2537$}, 11)
};
\addplot+[styleRL, bar shift=+4pt, forget plot, fill opacity=0.33] coordinates {
  ({$|A|\,{=}\,100$\\$|V_\mathrm{tile}|\,{=}\,598 $}, 10)
  ({$|A|\,{=}\,200$\\$|V_\mathrm{tile}|\,{=}\,975 $}, 16)
  ({$|A|\,{=}\,300$\\$|V_\mathrm{tile}|\,{=}\,1333$}, 23)
  ({$|A|\,{=}\,500$\\$|V_\mathrm{tile}|\,{=}\,2537$}, 44)
};
\resetstackedplots
\addplot+[styleILP, bar shift=+12pt] coordinates {
  ({$|A|\,{=}\,100$\\$|V_\mathrm{tile}|\,{=}\,598 $}, 16)
  ({$|A|\,{=}\,200$\\$|V_\mathrm{tile}|\,{=}\,975 $}, 60)
  ({$|A|\,{=}\,300$\\$|V_\mathrm{tile}|\,{=}\,1333$}, 115)
  ({$|A|\,{=}\,500$\\$|V_\mathrm{tile}|\,{=}\,2537$}, 314)
};
\legend{\ALGG, \ALGLRR, \ALGRL, \ALGILP}
\end{groupplot}
\end{tikzpicture}
\end{adjustbox}
\caption{
\rev{
Average initial and lifelong per-step computation times. For our method, the darkest shade represents the guidance step, the medium shade the rebalancing step, and the lightest shade the matching step.
}
\label{fig:cpu_time}
}
\end{figure}
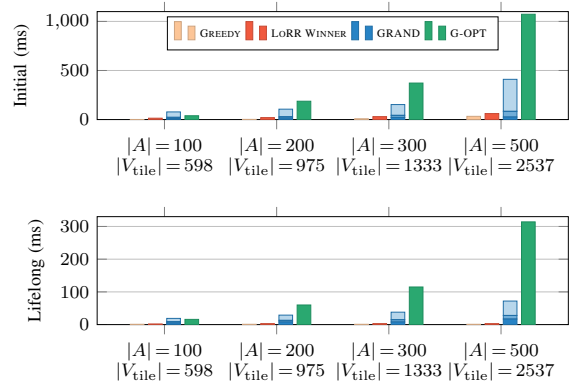

%% file: figures/plot_generalize_agent.tex
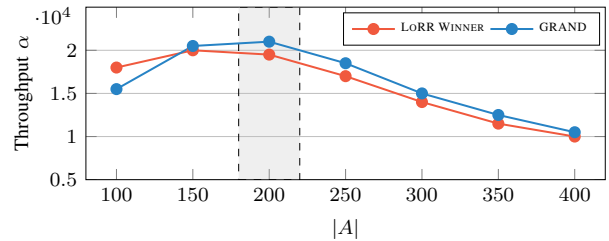
\begin{figure}[b]
\setlength{\abovecaptionskip}{-2pt}
\begin{adjustbox}{max width=\Scale\columnwidth, center}
\begin{tikzpicture}
\begin{axis}[
  width=\columnwidth, height=4.0cm, 
  xlabel={$|A|$},
  xlabel style={font=\footnotesize},
  ylabel={Throughput~$\alpha$},
  ylabel style={font=\footnotesize},
  yticklabel style={font=\footnotesize},
  xticklabel style={font=\footnotesize},
  xmin=80, xmax=420,
  ymin=5000,  ymax=25000,
  xtick={100,150,200,250,300,350,400},
  ytick={5000, 10000, 15000, 20000},
  ymajorgrids=true,
  y tick scale label style={at={(-0.1,0.9)}, anchor=south west},
  legend style={at={(0.99,0.97)}, anchor=north east, legend columns=2, font=\tiny},
  legend cell align=left,
  cycle list={{orange, mark=*, thick}, {green!60!black, mark=*, thick}},
]

\addplot[
  draw=black, dashed,
  fill=gray!50, fill opacity=0.25,
  forget plot
] coordinates {(180,5000) (220,5000) (220,25000) (180,25000) (180,5000)} -- cycle;

\addplot+[styleLRR2] coordinates
  {(100,18000) (150,20000) (200,19500) (250,17000) (300,14000) (350,11500) (400,10000)};
\addlegendentry{\ALGLRR}

\addplot+[styleRL2] coordinates
  {(100,15500) (150,20500) (200,21000) (250,18500) (300,15000) (350,12500) (400,10500)};
\addlegendentry{\ALGRL}

\end{axis}
\end{tikzpicture}
\end{adjustbox}
\caption{Throughput for $|V_\mathrm{tile}| = 975$. Our method is trained on $|A|=200$ and deployed on the other instances without retraining.}
\label{fig:generalize_agent}
\end{figure}

%% file: figures/plot_generalize_size.tex
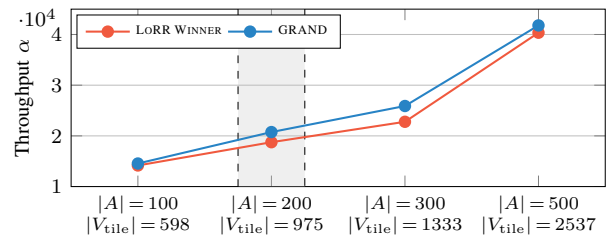
\begin{figure}
\setlength{\abovecaptionskip}{-2pt}
\begin{adjustbox}{max width=\Scale\columnwidth, center}
\begin{tikzpicture}
\begin{axis}[
  width=\columnwidth, height=4.0cm, 
  ylabel={Throughput~$\alpha$},
  ylabel style={font=\footnotesize},
  yticklabel style={font=\footnotesize},
  xmin=0.5, xmax=4.5,
  ymin=10000,  ymax=45000,
  xtick={1,2,3,4},
  xticklabels={
    {$|A|\,{=}\,100$\\$|V_\mathrm{tile}|\,{=}\,598 $},
    {$|A|\,{=}\,200$\\$|V_\mathrm{tile}|\,{=}\,975 $},
    {$|A|\,{=}\,300$\\$|V_\mathrm{tile}|\,{=}\,1333$},
    {$|A|\,{=}\,500$\\$|V_\mathrm{tile}|\,{=}\,2537$}
  },
  xticklabel style={align=center, font=\scriptsize},
  ymajorgrids=true,
  y tick scale label style={at={(-0.1,0.9)}, anchor=south west},
  legend style={at={(0.01,0.97)}, anchor=north west, legend columns=2, font=\tiny},
  legend cell align=left,
  cycle list={{orange, mark=*, thick}, {green!60!black, mark=*, thick}},
]

\addplot[
  draw=black, dashed,
  fill=gray!50, fill opacity=0.25,
  forget plot
] coordinates {(1.75,5000) (2.25,5000) (2.25,50000) (1.75,50000) (1.75,5000)} -- cycle;

\addplot+[styleLRR2] coordinates
  {(1,14164) (2,18752) (3,22781) (4,40350)};
\addlegendentry{\ALGLRR}

\addplot+[styleRL2] coordinates
  {(1,14549) (2,20752) (3,25894) (4,41788)};
\addlegendentry{\ALGRL}

\end{axis}
\end{tikzpicture}
\end{adjustbox}
\caption{\rev{Throughput for different scales. Our method is trained on the shaded instance and deployed on the other instances without retraining.}}
\label{fig:generalize_size}
\end{figure}

%% file: figures/table_ablation.tex
\begin{table}
\setlength{\tabcolsep}{4.75pt}
\setlength{\abovecaptionskip}{0pt}
\centering
\footnotesize
\renewcommand{\arraystretch}{\TableVerStretch} 
\rev{
\caption{Ablation of the different steps in our method. Evaluation with 200 agents and a map size of 975 without reassignment.}
\begin{adjustbox}{max width=\Scale\columnwidth, center}
\begin{NiceTabular}{c|c|ccc|c}
    \toprule
    \Block[c]{2-1}{\textbf{Metric}} & \Block[c]{2-1}{\textbf{\ALGRL}} & \Block[c]{1-3}{\underline{Guidance}} & & & \underline{Matching} \\
    & & Uniform & Demand  & Random & Greedy\\
    \midrule
    \textbf{Throughput} & \textbf{17916} & 16433 & 14910 & 14281 & 14476 \\
    \textbf{Time-to-task} & \textbf{27.1 s} & 35.5 s & 32.7 s & 38.5 s & 69.3 s \\
    \textbf{Time-in-task} & \textbf{83.9 s} & 85.8 s & 101.1 s & 101.6 s & 68.4 s \\
    \bottomrule
\end{NiceTabular}
\end{adjustbox}
\label{table:ablation}
}
\end{table}

%% file: sections/08-conclusion.tex
\section{Conclusion}

We proposed a hybrid task–scheduling architecture that pairs a learned global guidance policy with a lightweight optimization layer for 
agent–task assignment. 
This separation yields higher throughput across scales, with gains attributable to reduced congestion, while keeping per-step latency
within a $1\,\mathrm{s}$ control budget and below the global matching baseline \ALGILP. 
The policy also transfers zero-shot across occupancy ratios and map sizes without tuning. 
The approach highlights the value of graph neural representations for warehouse-like, graph-structured environments, providing compact and transferable inductive bias for large fleets and maps.

\paragraph*{Limitations and outlook}
Key directions include: (i) co-design with path planning via joint learning to further reduce congestion; \rev{(ii) closing sim-to-real gaps via log-driven calibration and 
better 
disturbance modeling, with explicit safety constraints 
for deploying 
GRAND on real systems;} (iii) scaling to heterogeneous agents/tasks, time windows, and priorities, with parallelism/accelerators for guidance and local assignment; and \rev{(iv) developing theory on stability, sample efficiency, and performance bounds for the hybrid \gls{acr:rl}–optimization scheme.}

In summary, learning-based global guidance coupled with optimization-based local assignment offers a practical, scalable blueprint for real-time, high-throughput scheduling in classical and lifelong \gls{acr:mapf}/\gls{acr:mapd}, and a foundation for co-designed task–motion systems at industrial scale.

%% file: references.bib
@article{zardini2022analysis,
  title={{Analysis and Control of Autonomous Mobility-on-Demand Systems}},
  author={Zardini, Gioele and Lanzetti, Nicolas and Pavone, Marco and Frazzoli, Emilio},
  journal={Annual Review of Control, Robotics, and Autonomous Systems},
  volume={5},
  number={1},
  pages={633--658},
  year={2022},
  publisher={Annual Reviews}
}

@article{d2012guest,
  title={{Guest Editorial: A Revolution in the Warehouse: A Retrospective on Kiva Systems and the Grand Challenges Ahead}},
  author={D'Andrea, Raffaello},
  journal={IEEE Transactions on Automation Science and Engineering},
  volume={9},
  number={4},
  pages={638--639},
  year={2012},
  publisher={IEEE}
}

@misc{waymo2025,
    title = {{Waymo Reports 250,000 Paid Robotaxi Rides per Week in U.S.}},
    note = {https://www.cnbc.com/2025/04/24/waymo-reports-250000-paid-robotaxi-rides-per-week-in-us.html},
    year = {2025},
    journal = {CNBC},
    author = {Elias, Jennifer and Kolodny, Lora}
}

@misc{amazon2025,
    title = {{Amazon Launches a New AI Foundation Model to Power its Robotic Fleet and Deploys its 1 Millionth Robot}},
    note = {https://www.aboutamazon.com/news/operations/amazon-million-robots-ai-foundation-model},
    year = {2025},
    author = {Dresser, Scott}
}

@inproceedings{ma2017lifelong,
  title={{Lifelong Multi-Agent Path Finding for Online Pickup and Delivery Tasks}},
  author={Ma, Hang and Li, Jiaoyang and Kumar, TK and Koenig, Sven},
  booktitle={Proceedings of the 16th Conference on Autonomous Agents and MultiAgent Systems},
  year={2017}
}

@inproceedings{yu2013structure,
  title={{Structure and Intractability of Optimal Multi-Robot Path Planning on Graphs }},
  author={Yu, Jingjin and LaValle, Steven},
  booktitle={Proceedings of the AAAI Conference on Artificial Intelligence},
  volume={27},
  pages={1443--1449},
  year={2013}
}

@article{doring2025parameterized,
  title={{Parameterized Complexity of Vehicle Routing}},
  author={D{\"o}ring, Michelle and Fehse, Jan and Friedrich, Tobias and Marten, Paula and Mohrin, Niklas and Simonov, Kirill and Soheil, Farehe and Timm, Jakob and Verma, Shaily},
  journal={arXiv preprint arXiv:2509.10361},
  year={2025}
}

@article{antonyshyn2023multiple,
  title={{Multiple Mobile Robot Task and Motion Planning: A Survey}},
  author={Antonyshyn, Luke and Silveira, Jefferson and Givigi, Sidney and Marshall, Joshua},
  journal={ACM Computing Surveys},
  volume={55},
  number={10},
  pages={1--35},
  year={2023},
  publisher={ACM New York, NY}
}

@article{hart1968formal,
  title={{A Formal Basis for the Heuristic Determination of Minimum Cost Paths}},
  author={Hart, Peter E and Nilsson, Nils J and Raphael, Bertram},
  journal={IEEE Transactions on Systems Science and Cybernetics},
  volume={4},
  number={2},
  pages={100--107},
  year={1968},
  publisher={IEEE}
}

@article{ramshaw2012minimum,
  title={{On Minimum-Cost Assignments in Unbalanced Bipartite Graphs}},
  author={Ramshaw, Lyle and Tarjan, Robert E},
  journal={HP Labs, Palo Alto, CA, USA, Tech. Rep. HPL-2012-40R1},
  volume={20},
  pages={14},
  year={2012}
}

@inproceedings{xu2022multi,
  title={{Multi-Goal Multi-Agent Pickup and Delivery}},
  author={Xu, Qinghong and Li, Jiaoyang and Koenig, Sven and Ma, Hang},
  booktitle={2022 IEEE/RSJ International Conference on Intelligent Robots and Systems (IROS)},
  pages={9964--9971},
  year={2022},
  organization={IEEE}
}

@inproceedings{kou2020idle,
  title={{Idle Time Optimization for Target Assignment and Path Finding in Sortation Centers}},
  author={Kou, Ngai Meng and Peng, Cheng and Ma, Hang and Kumar, TK Satish and Koenig, Sven},
  booktitle={Proceedings of the AAAI Conference on Artificial Intelligence},
  volume={34},
  pages={9925--9932},
  year={2020}
}

@article{wang2025paths,
  title={{Where Paths Collide: A Comprehensive Survey of Classic and Learning-Based Multi-Agent Pathfinding}},
  author={Wang, Shiyue and Xu, Haozheng and Zhang, Yuhan and Lin, Jingran and Lu, Changhong and Wang, Xiangfeng and Li, Wenhao},
  journal={arXiv preprint arXiv:2505.19219},
  year={2025}
}

@article{chen2021integrated,
  title={{Integrated Task Assignment and Path Planning for Capacitated Multi-Agent Pickup and Delivery}},
  author={Chen, Zhe and Alonso-Mora, Javier and Bai, Xiaoshan and Harabor, Daniel D and Stuckey, Peter J},
  journal={IEEE Robotics and Automation Letters},
  volume={6},
  number={3},
  pages={5816--5823},
  year={2021},
  publisher={IEEE}
}

@inproceedings{makino2024online,
  title={{Online Multi-Agent Pickup and Delivery with Task Deadlines}},
  author={Makino, Hiroya and Ito, Seigo},
  booktitle={2024 IEEE/RSJ International Conference on Intelligent Robots and Systems (IROS)},
  pages={8428--8434},
  year={2024},
  organization={IEEE}
}

@article{li2025fico,
  title={{FICO: Finite-Horizon Closed-Loop Factorization for Unified Multi-Agent Path Finding}},
  author={Li, Jiarui and Zanardi, Alessandro and Zhang, Runyu and Zardini, Gioele},
  journal={arXiv preprint arXiv:2511.13961},
  year={2025}
}

@inproceedings{agrawal2023rtaw,
  title={{RTAW: An Attention Inspired Reinforcement Learning Method for Multi-Robot Task Allocation in Warehouse Environments}},
  author={Agrawal, Aakriti and Bedi, Amrit Singh and Manocha, Dinesh},
  booktitle={2023 IEEE International Conference on Robotics and Automation (ICRA)},
  pages={1393--1399},
  year={2023},
  organization={IEEE}
}

@article{wang2025breaking,
  title={{Breaking the Hierarchy: Taxonomies and Survey on Multi-robot Integrated Task and Motion Planning}},
  author={Wang, Hanfu and Ye, Weibin and Wang, Jingchuan and Chen, Weidong},
  journal={techrxiv preprint techrxiv.173933261.11143776},
  year={2025},
}

@article{alonso2017demand,
  title={{On-Demand High-Capacity Ride-Sharing via Dynamic Trip-Vehicle Assignment}},
  author={Alonso-Mora, Javier and Samaranayake, Samitha and Wallar, Alex and Frazzoli, Emilio and Rus, Daniela},
  journal={Proceedings of the National Academy of Sciences},
  volume={114},
  number={3},
  pages={462--467},
  year={2017},
  publisher={National Academy of Sciences}
}

@inproceedings{gammelli2021graph,
  title={{Graph Neural Network Reinforcement Learning for Autonomous Mobility-on-Demand Systems}},
  author={Gammelli, Daniele and Yang, Kaidi and Harrison, James and Rodrigues, Filipe and Pereira, Francisco C and Pavone, Marco},
  booktitle={2021 60th IEEE Conference on Decision and Control (CDC)},
  pages={2996--3003},
  year={2021},
  organization={IEEE}
}

@inproceedings{gammelli2023graph,
  title={{Graph Reinforcement Learning for Network Control via Bi-Level Optimization}},
  author={Gammelli, Daniele and Harrison, James and Yang, Kaidi and Pavone, Marco and Rodrigues, Filipe and Pereira, Francisco C},
  booktitle={International Conference on Machine Learning},
  pages={10587--10610},
  year={2023},
  organization={PMLR}
}

@article{tresca2025robo,
  title={{Robo-taxi Fleet Coordination at Scale via Reinforcement Learning}},
  author={Tresca, Luigi and Schmidt, Carolin and Harrison, James and Rodrigues, Filipe and Zardini, Gioele and Gammelli, Daniele and Pavone, Marco},
  journal={arXiv preprint arXiv:2504.06125},
  year={2025}
}

@inproceedings{haarnoja2018soft,
  title={{Soft Actor-Critic: Off-Policy Maximum Entropy Deep Reinforcement Learning with a Stochastic Actor}},
  author={Haarnoja, Tuomas and Zhou, Aurick and Abbeel, Pieter and Levine, Sergey},
  booktitle={International Conference on Machine Learning},
  pages={1861--1870},
  year={2018},
  organization={PMLR}
}

@inproceedings{chan2024league,
  title={{The League of Robot Runners Competition: Goals, Designs, and Implementation}},
  author={Chan, Shao-Hung and Chen, Zhe and Guo, Teng and Zhang, Han and Zhang, Yue and Harabor, Daniel and Koenig, Sven and Wu, Cathy and Yu, Jingjin},
  booktitle={ICAPS 2024 System's Demonstration track},
  year={2024}
}

@article{yukhnevich2025enhancing,
  title={{Enhancing PIBT via Multi-Action Operations}},
  author={Yukhnevich, Egor and Andreychuk, Anton},
  journal={arXiv preprint arXiv:2511.09193},
  year={2025}
}

@article{gao2025adaptive,
  title={{Adaptive Congestion-Based Algorithms for Multi-Goal Task Assignment and Path Finding in Large-Scale Multi-Agent Systems}},
  author={Gao, Ye and Ding, Hang and Wang, Yuxuan and Zhang, Junjie and Sun, Qian and Zhang, Qian and Huang, Yiwen and Luo, Mao and Su, Zhouxing and Ding10, Junwen and others},
  journal={The League of Robot Runners Virtual Expo 2025},
  year={2025}
}

@inproceedings{chen2024traffic,
  title={{Traffic Flow Optimisation for Lifelong Multi-Agent Path Finding}},
  author={Chen, Zhe and Harabor, Daniel and Li, Jiaoyang and Stuckey, Peter J},
  booktitle={Proceedings of the AAAI Conference on Artificial Intelligence},
  volume={38},
  number={18},
  pages={20674--20682},
  year={2024}
}

@inproceedings{zang2025online,
  title={{Online Guidance Graph Optimization for Lifelong Multi-Agent Path Finding}},
  author={Zang, Hongzhi and Zhang, Yulun and Jiang, He and Chen, Zhe and Harabor, Daniel and Stuckey, Peter J and Li, Jiaoyang},
  booktitle={Proceedings of the AAAI Conference on Artificial Intelligence},
  volume={39},
  pages={14726--14735},
  year={2025}
}

@article{kovacs2015minimum,
  title={{Minimum-Cost Flow Algorithms: An Experimental Evaluation}},
  author={Kov{\'a}cs, P{\'e}ter},
  journal={Optimization Methods and Software},
  volume={30},
  number={1},
  pages={94--127},
  year={2015},
  publisher={Taylor \& Francis}
}

@book{bertsekas1998network,
  title={{Network Optimization: Continuous and Discrete Models}},
  author={Bertsekas, Dimitri},
  volume={8},
  year={1998},
  publisher={Athena Scientific}
}

@inproceedings{shi2021masked,
  title={{Masked Label Prediction: Unified Message Passing Model for Semi-Supervised Classification}},
  author={Shi, Yunsheng and Huang, Zhengjie and Feng, Shikun and Zhong, Hui and Wang, Wenjin and Sun, Yu},
  booktitle={Proceedings of the Thirtieth International Joint Conference on Artificial Intelligence},
  year={2021}
}
